\documentclass[a4paper,11pt,english]{article}

\usepackage{arxiv}
\renewcommand{\shorttitle}{\normalsize Semi-Supervised Bayesian GANs with Log-Signatures for \\ Uncertainty-Aware Credit Card Fraud Detection}

\usepackage[english]{babel}
\usepackage{subfig} 
\usepackage{xcolor}
\usepackage{amsmath}
\usepackage{amssymb}
\usepackage[utf8]{inputenc}
\usepackage{graphicx}
\usepackage{lmodern}
\usepackage{fancyvrb}
\usepackage{listings}
\usepackage{color}
\usepackage{tikz}
\usepackage{rotating}
\usepackage{multirow}
\usepackage{makecell}
\usepackage{array}
\usepackage{csquotes}
\usepackage{hyperref}
\usepackage{algorithm}
\usepackage{algorithmic}
\usepackage{float}
\usepackage{filecontents}
\usepackage{footnote}
\usepackage{chngcntr}
\usepackage{mathtools}
\usepackage{hhline}
\usepackage[english, noabbrev,capitalize,nameinlink]{cleveref} 
\usepackage{booktabs}
\usepackage{dsfont}

\usepackage[numbers,sort]{natbib}

\definecolor{denim}{rgb}{0.08, 0.38, 0.74}
\hypersetup{
	linktocpage=false,
	bookmarks=true,         
	unicode=false,          
	pdftoolbar=true,        
	pdfmenubar=true,        
	pdffitwindow=false,     
	pdfstartview={FitH},    
	pdftitle={Synthetic financial data},    
	pdfauthor={David Hirnschall},     
	pdfsubject={Research Proposal},   
	pdfcreator={David Hirnschall},   
	pdfproducer={David Hirnschall}, 
	pdfkeywords= {GANs} {Signatures} , 
	pdfnewwindow=true,      
	colorlinks=true,       
	linkcolor=black,   
	citecolor=black,    
	filecolor=black,   
	urlcolor=black,    
}

\newcolumntype{C}[1]{>{\centering\arraybackslash}m{#1}}

\definecolor{mygray}{rgb}{0.5,0.5,0.5}

\crefname{lstlisting}{Listing}{Listings}
\crefname{equation}{Equation}{equations}
\crefname{line}{Algorithm}{algorithms}

\usepackage{authblk}
\linethickness{2pt}
\title{\Large Semi-Supervised Bayesian GANs with Log-Signatures for Uncertainty-Aware Credit Card Fraud Detection}
\author{\textbf{David Hirnschall}}
\affil{Institute for Statistics and Mathematics\hfill\\ 
	Vienna University of Economics and Business \hfill\\
	Welthandelsplatz 1, 1020 Vienna, Austria \hfill\\
	\texttt{david.hirnschall@wu.ac.at}}
\date{}   

\usepackage{abstract}

\newtheorem{definition}{Definition}

\usepackage{diagbox}

\begin{document}
	\maketitle
	\begin{abstract}
		We present a novel deep generative semi-supervised framework for credit card fraud detection, formulated as time series classification task. As financial transaction data streams grow in scale and complexity, traditional methods often require large labeled datasets, struggle with time series of irregular sampling frequencies and varying sequence lengths. To address these challenges, we extend conditional Generative Adversarial Networks (GANs) for targeted data augmentation, integrate Bayesian inference to obtain predictive distributions and quantify uncertainty, and leverage log-signatures for robust feature encoding of transaction histories. We introduce a novel Wasserstein distance-based loss to align generated and real unlabeled samples while simultaneously maximizing classification accuracy on labeled data. Our approach is evaluated on the BankSim dataset, a widely used simulator for credit card transaction data, under varying proportions of labeled samples, demonstrating consistent improvements over benchmarks in both global statistical and domain-specific metrics. These findings highlight the effectiveness of GAN-driven semi-supervised learning with log-signatures for irregularly sampled time series and emphasize the importance of uncertainty-aware predictions.
	\end{abstract}
	
	
\section{Introduction}
As technology evolves rapidly and the volume of financial transactions surges, digital banking is experiencing unprecedented growth. Consequently, financial systems are increasingly vulnerable to fraudulent activities, such as credit card or online payment fraud \cite{dalpoz2017credit, jurgovsky2018sequence, 10341223}. Reliable fraud detection has therefore become a priority for financial institutions. Traditional methods, often rely on point estimates and require large amounts of labeled data \cite{10595068}. Class imbalance is mainly approached through oversampling and undersampling techniques \cite{fiore2019using, ijfs11030110, zhao2024resampling}. 

Existing methods can be broadly divided into two groups: classifiers using single-transaction features or sequential models that incorporate historical customer behavior \cite{raval2023raksha, benchaji2021enhanced}. The former often lack robustness, since what constitutes suspicious activities for one customer may be entirely normal for another. Sequential models usually rely on fixed-size time windows of transaction histories. Such windows may truncate long-term dependencies or, in case of new accounts, include potentially misleading transactions from other customers. A more natural approach is to include each individual customer's full transaction history. However, these transaction time series are usually irregularly sampled and vary in length, which cannot be handled easily by existing approaches, and highlights the need for effective feature extraction techniques. 

Handcrafted features, typically derived from expert knowledge or general statistics, such as mean, variance, skewness, and kurtosis, have been widely studied and can provide insights (see e.g.  \cite{jurgovsky2018sequence, bahnsen2016feature}). However, they generally fail to capture the temporal order of transactions, which is essential in anomaly detection for time series. To address these limitations, we propose encoding transaction histories using (log-)signatures. Signatures provide a mathematical representation of a path, gradually encoding details until fully characterizing it under mild conditions. They are robust to irregular sampling frequencies and varying time series lengths \cite{morrill2020generalised}. 

Classical machine learning models for time series classification (TSC) \cite{faouzi2022time} typically rely on large labeled datasets, which are often costly to acquire. Semi-supervised learning (SSL) mitigates this by leveraging abundant unlabeled data to learn representations that capture shared structures and reduce dependence on scarce labeled data \cite{springenberg2015unsupervised}. In practice, however, fraud datasets are constrained not only by the lack of labels but also by a shortage of unlabeled samples, in addition to severe class imbalance. Consequently, data augmentation became an essential component of regularization-based SSL and classification tasks \cite{ijfs11030110}. Inspired by successes in computer vision and natural language processing \cite{van2020survey}, we extend generative SSL techniques to sequential financial data. As common augmentation methods, such as rotation and cropping, may disrupt temporal dependencies, we employ generative models for data augmentation. 

Another key challenge in fraud detection is evaluation. Global measures such as ROC AUC or PR AUC provide useful overall summaries, but they do not reflect operational constraints, where only a very small fraction of transactions can be manually reviewed. To address this, we introduce a dual evaluation framework combining standard global metrics with domain-specific head metrics.  In particular, we evaluate based on Precision@K, Recall@K, and a cost-sensitive Expected Cost@K measure that incorporates the monetary impact of missed frauds and false alerts. This enables comparison not only in terms of statistical performance but also in terms of real-world business impact.

Further, to quantify uncertainty in predictions, a vital aspect in high-risk areas such as fraud-detection, we propose using a Bayesian inference. In a Bayesian setting, a distribution is placed over network weights, resulting in a predictive distribution rather than point estimates. This allows to predict fraud likelihood prediction while enabling uncertainty quantification.  Such uncertainty is particularly valuable, as point estimates may appear highly confident, yet the underlying distribution can reveal substantial variance. By modeling this uncertainty, our approach provides calibrated confidence information and increases robustness in the top $K\%$ ranking of fraud predictions.

In this paper, we integrate recent advances in generative semi-supervised learning from computer vision to developments in credit card fraud detection. Specifically, extend GAN-based SSL from the image domain to sequential transaction data. Our main contributions are as follows:
\begin{itemize}
	\item We introduce a novel loss to unify input dimensions, effectively capture temporal dependencies in transaction sequences, and minimize discrepancies between generated and real unlabeled samples, while maximizing classification accuracy on labeled data. This loss is based on the Wasserstein distance combined with log-signatures.
	\item We introduce a conditional generator to produce tailored augmentations by controlling categorical feature combinations (e.g., customer age or risk group), ensuring realistic and context-aware synthetic samples.
	\item We enhance generalization and robustness by integrating Bayesian inference over network weights, thereby providing predictive distributions and principled uncertainty estimates for fraud detection.
	\item We provide a comprehensive dual evaluation framework combining statistical and domain-specific, cost-sensitive metrics, ensuring practically relevant performance assessment. 
\end{itemize}
Our approach is validated on the BankSim dataset \cite{LopezRojas2014} under varying proportions of labeled data. As access to real-world financial data is extremely restricted, due to privacy and regulatory concerns, many studies rely on simulated data. In particular, BankSim is an agent-based simulator designed to mimic real banking behavior and has become a widely adopted benchmark in this domain. Results demonstrate improvements over benchmark models in terms of both global and domain-specific head metrics, while providing distributional outputs and uncertainty measures. \\
The remainder of this paper is organized as follows. \cref{recentwork} reviews related literature on SSL and path signatures for time series modeling. \cref{preliminary} briefly recaps necessary background on GANs, Bayesian neural networks and log signature. \cref{model} introduces the proposed network architecture, loss functions and training procedure for our approach. \cref{results} describes the BankSim dataset, outlines the evaluation process and presents the numerical results. Section \cref{conclusion} draws the conclusions and discusses directions for future work.

\section{Literature review}\label{recentwork}
\textbf{Semi-supervised Learning (SSL):}
Methods for extracting additional information from unlabeled data range from deep label propagation \cite{iscen2019label} to more recent approaches \cite{berthelot2019mixmatch}, unifying regularization with pseudo-labels into a single framework. Despite their successes, most of these methods were not initially designed for the TSC tasks and therefore ignore temporal relations \cite{van2020survey}.

\textbf{SSL for Time Series:}
For time series data, SSL approaches typically fall into two categories: self-learning methods, where unlabeled samples are iteratively assigned pseudo-labels, and regularization-based methods that exploit shared structures across labeled and unlabeled data. Recent work explores enhanced data augmentation strategies \cite{goschenhofer2022deep} or training a model jointly for supervised classification on labeled data and an auxiliary forecasting task on all samples \cite{jawed2020self}. GAN-based techniques extend this line by regenerating signals and combining unsupervised representation learning with supervised loss components \cite{rezagholiradeh2018reg}.

\textbf{GAN-based SSL:}
\cite{salimans2016improved} propose labeling generated samples as new $(K+1)^{th}$ class and solving a $(K+1)$ class classification problem using a GAN. 
\cite{liu2020does} provide theoretical insights on suboptimal generators potentially improving SSL by moving the discriminator's decision boundaries to high-density areas of the data manifold. As in an SSL framework, the discriminator might perform well, whereas the generator might still produce visually unrealistic samples \cite{li2017triple} introduced Triple GAN, containing three neural networks. Three-player GANs for missing value imputation were proposed by \cite{pan2024segansemisupervisedlearningapproach}. \cite{toutouh2023semi} provide a comprehensive overview of SSL-GAN training enhancements and propose semi-supervised GANs with spatial coevolution for image datasets. \cite{kralik2024gan} used a WGAN-based semi-supervised approach for anomaly detection.

\textbf{Signatures for feature engineering:} 
Signatures, first introduced by \cite{chen1954iterated} in the 50s, became widely recognized in the mathematical community through Terry Lyons' development of Rough Path theory \cite{lyons1998differential}. More recently, they gained considerable traction in the machine learning community. In particular (log-)signatures have emerged as non-parametric and mathematically principled dimension reduction technique for time sereis data \cite{sturm2025path}, which has led to successful applications across a broad range of domains, including pricing derivatives \cite{lyons2019numerical}, human action and gesture recognition \cite{yang2022developing, gesture_sig} and more recently sports analytics \cite{hirnschall2025path}. In financial time series encoding and generative modeling, \cite{buehler2020data} developed a market simulator trained on path signatures and \cite{liao2019learning} combined log-signatures with recurrent neural networks to learn neural stochastic differential equations. Other works exploit signatures to measure time series similarities \cite{ni2021sig}, detect market anomalies \cite{akyildirim2022applications} and enhance deep learning architectures, such as transformers for time series modeling and deep hedging \cite{tong2023sigformer, moreno-pino2024rough}.

\section{Preliminary}\label{preliminary}
\subsection{Problem Setting}\label{problem}
We formulate the fraud detection problem as a general classification task with $K$ classes and two available data sets $\mathcal{D}_{ul}=\{x^{(i)}\}_{i=1}^N$ of $N$ unlabeled samples of multivariate time series, including continuous and categorical features and $\mathcal{D}_l=\{x^{(i)}_k, y^{(i)}_k\}_{i=1}^{N_l}$ a set of $N_l$ samples with the corresponding class labels 
$y_k^{(i)}$ where $k=1,\dots,K$ indicates the class and $N_l\ll N$ holds. The goal of conditional GAN-based semi-supervised learning is to simultaneously train a generative model $G$ to simulate samples $G(z,cond)$ given a latent input $z$ and some condition $cond$. Those samples are used to train a classification model $D$ on all available data $\mathcal{D}_l\cup \mathcal{D}_{ul} \cup \{x|x=G(z,cond)\}$, exploiting generative representation of the data to improve classification performance beyond what could be achieved using labeled data alone.

\subsection{Wasserstein Generative Adversarial Networks (WGANs)}\label{gans}
Introduced by \cite{goodfellow2020generative} for image generation, GANs contain two neural networks playing a min-max game. They are trained simultaneously on each other's feedback. While the first model, referred to as generator $G$, is a map that transports a latent distribution $p(z)$ to a model distribution $p_{model}$ to best approximate the real data distribution $p_{data}$, the second model, called discriminator $D$, classifies whether a given sample is real or generated. 

Wasserstein GANs \cite{arjovsky2017wasserstein} use a continuous learning curve even for non-overlapping distributions. Therefore, the distance between distributions $\mu$ and $\nu$ is measured by the Wasserstein metric,
\[W_1(\mu, \nu)=\inf _{\gamma \in \Pi\left(\mu, \nu\right)} \mathbb{E}_{(X, Y) \sim \gamma}[d(X,Y)],\]
where $\Pi\left(\mu, \nu\right)$ denotes the set of all joint distributions with marginals $\mu$ and $\nu.$ Joint distributions can not be observed from market data, hence the Kantorovich-Rubinstein dual representation 
\[W_1\left(\mu, \nu\right)=\sup _{\|f\|_{L} \leq 1} \mathbb{E}_{X \sim \mu}[f(X)]-\mathbb{E}_{Y \sim \nu}[f(Y)],\]
where $\|.\|_{L}$ denotes the Lipschitznorm, which is used for implementations. The test function $f$ is approximated by a neural network $D$. 
\noindent To enforce Lipschitz continuity of $D$, \cite{arjovsky2017wasserstein} proposed clipping the weights to a compactum $[-c, c]$.  \cite{gulrajani2017improved} added a gradient penalty term penalizing $D$ for gradients unequal to 1. 

\subsection{Bayesian Inference in GANs}\label{bayes}
Bayesian GAN were first introduced by \cite{bayesgan} to model uncertainty and mitigate model collapse by enhancing the diversity of generated data samples. They propose placing prior distributions $p(\theta_g, \alpha_g)$ and $p(\theta_d, \alpha_d)$ with parameters $\alpha_g$ and $\alpha_d$ over network weights $\theta_g$ and $\theta_d$ and utilizing the Stochastic Gradient Hamiltonian Monte Carlo (SGHMC) \cite{chen2014stochastic} algorithm to marginalize the corresponding posterior distributions. For a latent vector $z$ and an observed data sample $X$, they draw weight samples iteratively from the conditional posteriors by combining the network likelihood functions with chosen priors. During each iteration, batches of generated and unlabeled and all available labeled samples are used. \cite{bissiri2016general} proposed a general framework for updating beliefs on $\theta$ given information $x$, minimizing the expected loss of $l(x,\theta)$, rather than the traditional likelihood functions, as follows
\[\theta_0=\arg \inf _{\theta \in \Theta} \int l(\theta, x) \mathrm{d} F_0(x),\]
where $F_0(x)$ is a unknown distribution from which i.i.d. observation arise. For prior beliefs $\pi(\theta)$ and $x$ observed from $F_0$, and argues that \[p(\theta \mid x) \propto \exp \{-l(\theta, x)\} p(\theta),\]
a valid and coherent update to the posterior $p(\cdot \mid x).$

\subsection{Log-Signatures for Feature Encoding}\label{challenges}
Path signatures offer a unique and compact characterization of sequences while capturing their structural properties in a mathematically rigorous manner \cite{hambly2010uniqueness}. This property is particularly valuable when dealing with variable-length time series of irregularly sampled time intervals, as is typically the case for transaction data. 
First defined for continuous paths of bounded variation and later extended to discrete paths by linear interpolation \cite{chevyrev2016primer}, the signature transformation of 
a $d$-dimensional time series $x=(x_i)_{i=1,\dots,n}$ and its piece-wise linear interpolation $X=(X_t)_{t\in[t_1,t_n]}$ with $X_{t_i}=x_i$ for $t_1, \dots, t_n$ 
is defined as follows.
\begin{definition} 
	For a continuous path with finite variation $X: [t_1,t_n] \rightarrow \mathbb{R}^d$ from a compact time interval $[t_1,t_n]$ to $\mathbb{R}^d$, the signature is defined by,
	\[\operatorname{S}(X)_{[t_1,t_n]} =(1, S(X)^{(i)}_{[t_1,t_n]},  \dots,  \left.S(X)^{\left(i_1, \ldots, i_N\right)}_{[t_1,t_n]}\dots\right)_{i_1, \ldots, i_N=1}^d,\]
	where for any $\left(i_1, \ldots, i_k\right) \in\{1, \ldots, d\}^k$,
	\[S(X)^{\left(i_1, \ldots, i_k\right)}_{[t_1,t_n]}=\int_{t_1 \leq u_1<\cdots<u_k \leq t_n} \cdots \int \mathrm{d} X_{u_1}^{i_1}\dots \mathrm{d} X_{u_k}^{i_k}\in \mathbb{R} .\]
	The truncated signature of $X$ of degree $M$ is denoted as 
	$S_M\left(X\right)_{[t_1,t_n]}=\left(1, S(X)^{(1)}_{[t_1,t_n]}, \ldots, S(X)^{(M)}_{[t_1,t_n]}\right).$
	
\end{definition}
The error made by the truncation at level $M$ decays with factorial speed as $\mathcal{O}(1/M!)$; see \cite{liao2023sig}. Note that for piece-wise linear paths computation no longer requires integrals, but by Chen's identity, they can be constructed directly from contributions of the individual line segments.

Log-signatures are parsimonious representations of signatures, removing redundancies and, therefore, reducing the dimension compared to the signature \cite{ni2021sig}. According to the shuffle product \cite[Theorem 1.14]{chevyrev2016primer}, every polynomial function on signatures can be expressed as a linear combination of signature terms, which introduces repeated information. e.g. $S(X)_{s,t}^{i, i}=\frac{1}{2}\left(S(X)_{s,t}^i\right)^2.$ These redundancies are removed by the log-signature, retaining the same information in fewer terms.\\
To define the log-signature, we recap the definition of the logarithm $log(a)$ of an element $a$ in a Tensor algebra space. 
\begin{definition} 
	Let $a=(a_0, a_1, \dots,a_n)$ be an element in a tensor algebra $T((\mathbb{R}^d))$ such that $a_0=1$ and $t:=(a-1)$. Then, the logarithm is defined by,
	\[\log(a)=\log(t+1)=\sum_{n=1}^\infty \frac{(-1)^{n-1}}{n}t^{\otimes n}, \quad\forall a\in T((\mathbb{R}^d)).\]
\end{definition}
\begin{definition} 
	The log-signature of a path $X:[t_1,t_n]\to \mathbb{R}^d$, denoted as $LogSig(X)_{[t_1,t_n]}$, is defined as the logarithm of the signature. The truncated log-signature of degree $M$ is denoted by $LogSig_M(X)_{[t_1,t_n]}$.
\end{definition}
Log-signatures are robust to irregular sampling and uniquely determine the path up to tree-like equivalences \cite{hambly2010uniqueness}. A detailed discussion of the log-signature in machine learning, including its dimension reduction capabilities and suitable path augmentations to enrich the original path, is given by \cite{morrill2020generalised}. 

\section{Proposed Model}\label{model}
To recap, our proposed GAN-based SSL approach relies on three main ideas: 1) Constructing a conditional generator to simulate meaningful samples by controlling for categorical feature combinations. 2) Introducing a novel loss function based on the Wasserstein distance and log-signatures that unifies input dimension and efficiently extracts temporal features of time series data to simultaneously minimize the discrepancy between real and generated unlabeled samples and classify real samples as a supervised learning task. 3) Placing distributions over network weights to avoid model collapse, enhance generalization of the discriminator and estimate the probability of the target variable rather than a point estimate.

\subsection{Network architecture}

\begin{figure*}[ht]
	\centering
	\subfloat[Discriminator network]{\includegraphics[width=1\linewidth]{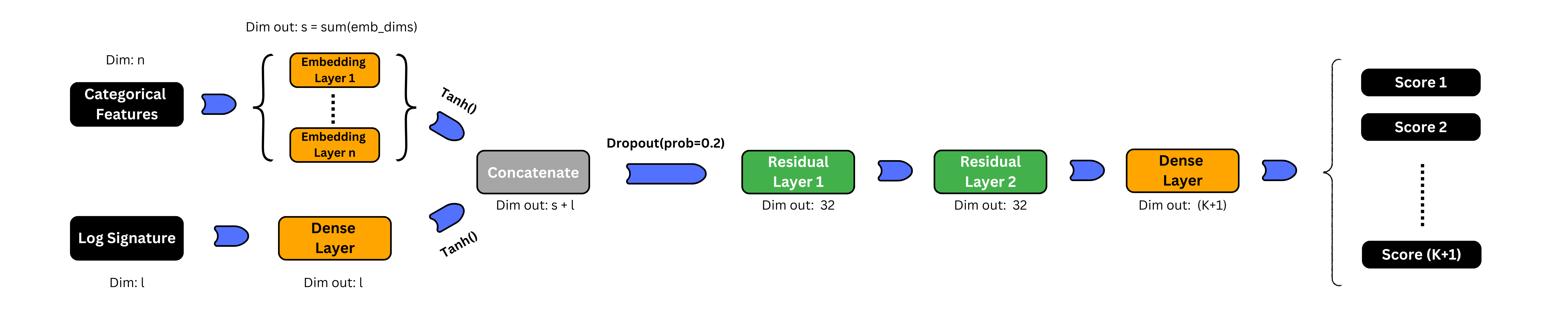}}	\\
	\subfloat[Generator network]{\includegraphics[width=1\linewidth]{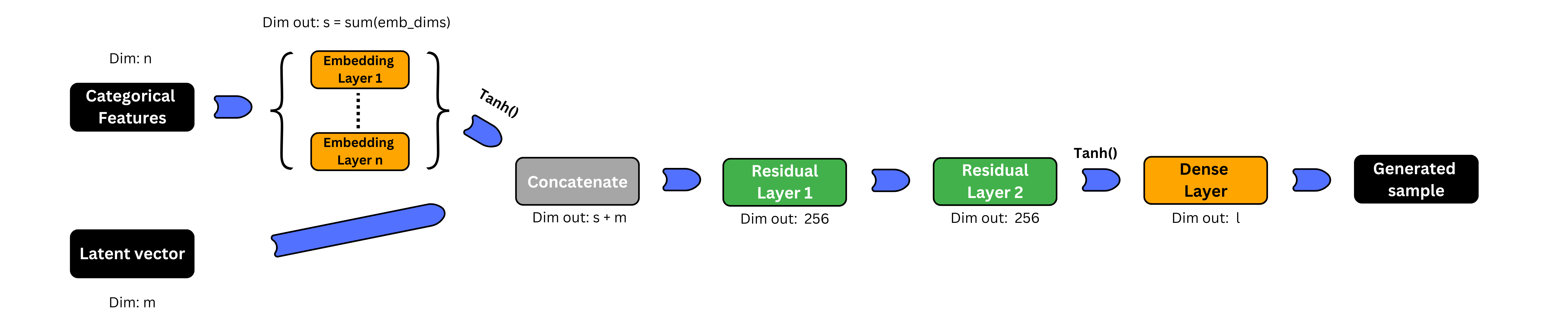}}
	\caption{Network architectures for discriminator and conditional generator.}
	\label{architectures}
\end{figure*}

\textbf{Generator:} For the generator, we generate samples directly representing log-signatures of augmented time series conditioned on a vector $cond$. To ensure a suitable combination of categorical values, $cond$ is sampled from real training data. 
Formally, we train a network $G$ that maps a latent vector $z$ and a vector $cond=(c_1,\dots, c_n)$ to an output $X_{fake}$ using tanh activation functions and residual layers defined as
\begin{definition}
	Let $F: \mathbb{R}^n \rightarrow \mathbb{R}^n$ be an affine transformation and $\phi$ a  tanh function. Then, a residual layer $R: \mathbb{R}^n \rightarrow \mathbb{R}^n$ is defined by
	\[R(x)=x+\phi \circ F(x),\]
	where $\phi$ is applied component-wise.
\end{definition}
First, each categorical feature is passed through an embedding layer, which maps it to a vector with dimension, $emb\_dims=(\#_{distinct}(c_1),\dots, \#_{distinct}(c_n) )$, determined by the number of distinct values of the feature in the training data followed by a tanh activation function. Second, we concatenate the output with the latent vector and apply two residual layers followed by a tanh and a fully connected layer, as illustrated in \cref{architectures}.

\textbf{Discriminator:} Similar to \cite{bayesgan}, we aim for a discriminator $D$ that takes into account class labels. We, however, propose a discriminator that returns a vector of raw scores with values in $\mathbb{R}^{(K+1)}$ instead of estimating the probability that sample $x^{(i)}$ belongs to class $y^{(i)}$, where class label 0 represents a sample produced by the generator. To do so, we construct a feedforward neural network $D$ using tanh activation functions and residual layers. \\
Further, if given a real sample, we add a preprocessing step to extract meaningful information from a given data sample $X$ containing both time series data $ts$ and a categorical feature vector $cond=(c_1,\dots, c_n)$. We first augment the time series using a time augmentation, a lead-lag augmentation and an invisibility-reset augmentation. Second, we apply a piece-wise linear interpolation and finally compute the truncated log-signature of order 4. The used augmentations ensure uniqueness of the log-signature, capture information on the quadratic variation of the process and add extra information about the starting point. The dimension of a $d$-dim time series increases to $2d+3$. A detailed overview of possible path augmentation and their classification is given by \cite{morrill2020generalised}. Finally, as in the generator, each categorical feature is passed through an embedding layer, mapping it to a vector with dimension, $emb\_dims=(\#_{distinct}(c_1),\dots, \#_{distinct}(c_n) )$. This is done for both real and synthetic samples.\\
For a given log signature of length $l$ and a vector $cond$ of $n$ categorical features, the discriminator network can be illustrated as \cref{architectures}.

\subsection{Loss functions}
Let $\{x^{(i)}\}_{i=1}^N$ be $N$ unlabeled observations and $\{x^{(i)}_k, y^{(i)}_k\}_{i=1}^{N_l}$ be $N_l$ labeled real observed with class labels $y^{(i)}_k\in \{1,\dots,K\}$. We label generated data as class 0. 
Given observed time series $X_{r}$, a latent vector $z$, and generated time series $X_{g}=G\left(z^{(i)} ; \theta_g\right)$, we apply the mentioned augmented and compute their truncated log-signatures.

While the generator only works with unlabeled data, the discriminator's loss is computed based on labeled and unlabeled data samples.

\textbf{Discriminator:} 
Let $d_r=(d_{r,0}, \dots, d_{r,K})$ and $d_g=(d_{g,0}, \dots, d_{g,K})$ be raw scores for $X_{r}$ and  $X_{g}$, respectively, as outputted by the discriminator D. The loss function contains three parts: one part for unlabeled data, one for labeled data and a gradient penalty term $GP$ to ensure Lipschitz continuity of the network \cite{gulrajani2017improved}. For the labeled samples, we compute a cross entropy loss,
\[L_{labeled}=-\frac{1}{N_l}\sum_{i=1}^{N_l}\sum_{k=1}^K y^{(i)}_k \log \left(\hat{y}^{(i)}_k\right),\]
with \[\hat y^{(i)}_k=\frac{\exp(d^{(i)}_{r,k})}{ \sum_{j=1}^K\exp(d^{(i)}_{r,j})}\] being the probability that sample $x^{(i)}$ belongs to class $y^{(i)}_k$. 
For the unlabeled data, we use a loss inspired by the Wasserstein loss function used in a WGAN. However, we do not approximate the test function $f$ directly by a neural network with an $\mathbb{R}$-dim output, but by the discriminator introduced above, followed by a function $T$ with 
\[T(x_0, \dots, x_{K})\to \frac{1}{\sqrt{(K+1)} }(x_0-\sum_{i=1}^Kx_i),\]
which is Lipschitz continuous with coefficient 1.
Note that this is equivalent to using a final linear output-layer with fixed weights $\frac{1}{\sqrt{(K+1)} }(1 ,-1 ,\dots,-1)$.
Intuitively, this pushes the network to give a high score to class label 0 and a low score to labels $\{1,\dots,K\}$ for synthetic samples, while doing the opposite for real samples. The unlabeled loss for a mini-batch of $n_d$ samples is defined as
\[L_{unlabeled}=
\frac1{n_d}\sum_{i=1}^{n_d} (T(d^{(i)}_r))- \frac1{n_d}\sum_{i=1}^{n_d}(T(d^{(i)}_g))
.\]
The gradient penalty term \cite{gulrajani2017improved} is computed for interpolated data samples following $\mathbb{P}_{i}$ as 
\[GP=\mathbb{E}_{\hat{X} \sim \mathbb{P}_{i}}\left[\left(\left\|\nabla D\left(\operatorname{LogSig}_M(\hat{X})\right)\right\|_2-1\right)^2\right].\]

Finally, for a latent input $z$, a categorical vector $cond$, a sample $X$, a scaling factor $\lambda$ and network weights $\theta_d$ and $\theta_g$ the loss function for the discriminator is defined as:
\begin{align*}
l\left(\theta_d \mid z, cond, X, \theta_g\right) &=L_{unlabeled}+\lambda L_{labeled}+ GP
\end{align*}
\textbf{Generator:} 
The generator's loss only depends on unlabeled data samples, as its goal is to generate samples that are classified as real by the discriminator. Hence, for a latent input $z$, a categorical vector $cond$ and network weights $\theta_d$ and $\theta_g$, the loss for a mini-batch of $n_g$ samples is
\[l\left(\theta_g \mid z, cond, \theta_d\right) =\frac1{n_g}\sum_{i=1}^{n_g}T(d_g^{(i)}).\]

\subsection{Posterior sampling}

To marginalize the posterior over the weights, we follow the general Bayesian updating approach (see \cref{bayes}) and use Stochastic Gradient Hamiltonian Monte Carlo (SGHMC), introduced in \cite{chen2014stochastic}. SGHMC is very closely related to momentum-based stochastic gradient descent (SGD). Hence, we can directly import parameter settings, such as learning rate and momentum terms. Empirically, however, we achieved better results using ADAM optimizer instead of momentum SGD. SGHMC extends the Hamiltonian gradient descent algorithm by using noisy gradients based on mini-batches of data, which allows the algorithm to scale as no gradients of big data batches have to be computed. Further, many practical benefits of Bayesian inference for GANs come from exploring a multimodal distribution of weights, which is enabled by SGHMC. 

Following \cite{bayesgan}, we set parameters $\alpha_g$ and $\alpha_d$ for the prior distributions $p\left(\theta_g \mid \alpha_g\right)$ and $p\left(\theta_d \mid \alpha_d\right)$ of the generator and discriminator weights, respectively and sample network initial weights $\left\{\theta_g^j\right\}_{j=1}^{MC_g}$ and $\left\{\theta_d^j\right\}_{j=1}^{MC_d}$ from the assumed prior distributions for $MC_g$ and $MC_d$ parallel running chains. We extract $n_g$ noise samples $\{z^{(i)}\}_{i=1}^{n_g}$ from the latent distribution $p(z)$ and draw a mini-batch of $n_d$ data samples $X_{real}=(ts,cond)$ of real data. Then, we use the SGHMC algorithm (see \cref{alg}) to update the parameters $\theta_g$ and $\theta_d$, respectively. Finally, the collected samples yield a predictive distribution and we use posterior mean as final prediction.

Note that clarity, we present one SGHMC iteration using a standard momentum-based SGD \cref{alg}, whereas, similar to \cite{bayesgan}, we achieved better results using ADAM optimization. Choosing a prior distribution is a crucial part of Bayesian inference. Hence, it often relies on expert knowledge. We avoid any exogenous assumptions or domain expert knowledge and follow the Glorot normal initialization \cite{glorot2010understanding} to maintain a high information flow across layers. Network weights are randomly drawn from the centered normal distribution
\[p(\theta_g) \sim \mathcal{N}\left(0, \sigma_{\text {prior }}^2 I_J\right),\]
with variance
\[\sigma^2_{\text {prior }}=g^2 \cdot \frac{2}{\text {fan in}+ \text {fan out}},\]
with scaling factor $g$ and fan in and fan out denoting the input and output dimensions of a network layer. As we use small layers down to 32 nodes, we choose $g=1$ to keep the variance reasonably small. Hyperparameters were adopted from commonly used settings in the literature, with minimal additional tuning of the learning rate. Based on this tuning study \cref{tab:lr_tuning}, we selected learning rates of 0.0001 for the generator and 0.005 for the discriminator in the final evaluation. The much higher learning rate for the discriminator D is based on the fact that it has to learn both classification and discrimination quickly, while the generator only learns through D's signal \cite{heusel2017gans}. Mini-batch size is chosen as 2048 and $\lambda=10.$ We train the generator/discriminator with alternating update steps using $n_{critic}=5$ discriminator updates per one generator update for 1000 epochs. 

\begin{algorithm}[h]
	\caption{One training epoch of SGD with friction term $\alpha$, learning rate $\eta$, $MC_g$ and $MC_{d}$ parallel running Markov Chains and previous posteriors samples $\left\{\theta_g^j\right\}_{j=1}^{MC_{g}}$ and $\left\{\theta_d^j\right\}_{j=1}^{MC_{d}}$.}
	\label{alg}
	\begin{algorithmic}
		
		\FOR{$j=1,\dots,MC_{g}$}
		\STATE - sample noise batch $\{z^{(m)}\}_{m=1}^{n_g}$ from $p(z)$ \\
		\STATE	- Update $\theta_g^{j}$ using SGHMC with $\epsilon\sim \mathcal{N}(0, 2 \alpha \eta I)$:
		\begin{align*}
		&\theta_g^{j} \leftarrow \theta_g^{j}+\nu ; \\&\nu \leftarrow(1-\alpha) \nu+\eta \frac{\partial \sum_{k=1}l\left(\theta_g^{j} \mid z, \theta_d^k\right)}{\partial \theta_g^{j}}+\epsilon
		\end{align*}
		\STATE - append $\theta_g^{j}$ to sample set.
		\ENDFOR
		\FOR{$j=1,\dots,MC_d$}
		\FOR{$i=1,\dots,n_{critic}$}
		\STATE - sample a batch of real data real from $P_{\text {data }}$: $X_{real}=\{X_{real}^{(1)}, \ldots, X_{real}^{(n_d)}\}$ \\
		
		\STATE- sample noise batch $z=\{z^{(m)}\}_{m=1}^{n_d}$ from $p(z)$.\\
		
		\STATE- Generate $X_{fake} = G(z)$, hence $\textbf{X} =X_{real} \cup X_{fake}$\\
		\STATE- Update $\theta_d^{j}$ using SGHMC with $\epsilon\sim \mathcal{N}(0, 2 \alpha \eta I)$:
		\begin{align*}
		&\theta_d^{j} \leftarrow \theta_d^{j}+\nu ; \\&\nu \leftarrow(1-\alpha) \nu+\eta \frac{\partial \sum_{k=1} l\left(\theta_d^{j} \mid z, \mathbf{X}, \theta_g^k\right)}{\partial \theta_d^{j}}+\epsilon 
		\end{align*}
		\STATE- append $\theta_d^{j}$ to sample set.
		\ENDFOR
		\ENDFOR

	\end{algorithmic}
\end{algorithm}

\section{Empirical Evaluation}\label{results}
\subsection{Dataset and Preprocessing}

The BankSim dataset is publicly available on Kaggle \cite{LopezRojas2014} and simulates financial transactions based on data provided by a bank in Spain over approximately six months. Because access to real banking records is restricted by privacy and regulatory constraints, BankSim has become a common benchmark in fraud detection research, providing a common ground for methodological comparison. While not including any personal or legally sensitive customer information, it effectively replicates realistic transaction patterns. As a result, it serves as a robust foundation for developing and evaluating fraud detection models, serving both academic researchers and practitioners. The authors simulated 594643 records, of which 7200 were fraudulent transactions. The dataset contains ten columns, including 7 categorical (\textit{customer, age, gender, zipcodeOri, merchant, zipMerchant} and \textit{category}), two continuous (\textit{step} and \textit{amount}) for each transaction and the target indicating if a transaction is fraudulent.

In this paper, we do not work directly with single transaction features but rather formulate the fraud detection problem as time series classification task. We, therefore, group transactions based on customer, resulting in a dataset of 4112 individual customers, of which 12 were excluded due to missing gender. Out of the remaining 4100 customers, 1479 contain fraudulent transactions. 
Each customer has conducted between 5 and 265 transactions of up to 8329.96 Euros. Since our sequential model requires a minimum transaction history before making predictions, we begin predicting after the first four transactions, ensuring each customer contributes at least one labeled sample. After an initial investigation, we observed that common transaction-level models such as Random Forest exhibit a dis-proportionally high dependence on the \textit{amount} feature. In fact, merely scaling this single variable substantially changed the results, indicating a lack of robustness to variations such as different currencies (see \cref{tab:scaling_effects}). To address this and ensure fair comparability, we trained all models on our prepared sequential data. This further allows each model to exploit the richer information contained in customers' transaction histories, rather than relying on isolated singe transactions.

We remove all identifiers for card holders and merchants to encourage generalization and only keep a small set of raw features comparable to previous work \cite{jurgovsky2018sequence, bahnsen2016feature}). We prepare the continuous features,
\begin{itemize}
	\item \textit{step difference}: Elapsed time since last transaction,
	\item \textit{amount}: Amount of money involved in the transaction,
\end{itemize}
and divide them by their maximum values, augment the path using a time augmentation, a lead-lag augmentation and an invisibility-reset augmentation and compute the log signature of order four of the complete customer history.
Further, we add the categorical features:
\begin{itemize}
	\item \textit{age}: Most recent age category of the customer,
	\item \textit{gender}: Most recent gender of the customer,
	\item \textit{risk level}: Risk level of the customer. Individual transactions are grouped into risk categories based on the percentage of fraudulent transactions within the corresponding category. These categories are defined by fraud rates in percent as [0,2], (2,10], (10,30], (30,50], (50,100]. A customer's risk level is determined as the weighted average of its transaction risk levels, weighted by the sequential order of the transactions.
\end{itemize}

\subsection{Baseline models}
Besides our proposed approach, we trained a diverse set of benchmark models for comparison. These include both traditional supervised machine learning classifiers and semi-supervised extensions. Following previous work in fraud detection \cite{makki2019experimental}, we selected seven representative baseline models covering linear, non-linear, tree-based and neural structures.

As classical benchmarks, we trained a Naive Bayes (NB), a logistic regression (LReg), a $k$-nearest neighbour (KNN) and a support vector machine (SVM). We further include a random forest as a widely used ensemble method in fraud detection and a fully supervised feed-forward neural network (FNN) as a deep learning model. The FNN is implemented with the same architecture as the discriminator described in \cref{architectures}, except that the final layer outputs two logits and the model is trained with standard cross-entropy loss. For a semi-supervised baseline, we include a self-training variant of logistic regression (LReg-SSL), utilizing the semi-supervised wrapper available in the scikit-learn Python package \cite{scikitlearn}. 

We do not include sequential models such as LSTM or transformer-based benchmark models. While widely applied to sequential data, they face well-known limitations when applied to irregularly sampled, variable-length histories \cite{lechner2020learninglongtermdependenciesirregularlysampled}. Their use would require ad hoc preprocessing, e.g., windowing, padding, time embeddings, or complex architectural adaptations that diverge from the core contribution of this work. Instead, we explicitly leave such extensions for future work and focus here on demonstrating that our Bayesian log-signature GAN naturally handles irregular sequences without imposing artificial structure.

\subsection{Evaluation Procedure}
Evaluating the performance of semi-supervised models requires special caution due to factors such as the selection of the labeled data points \cite{van2020survey}.
\cite{oliver2018realistic} provide guidelines for the realistic evaluation of semi-supervised models to guarantee unbiased and fair comparison results. This improved procedure includes splitting a fully labeled dataset $\mathcal{D}$ into a small labeled dataset $\mathcal{D}_l$ and a large unlabeled dataset $\mathcal{D}_{ul}$ with artificial unlabeling of randomly drawn samples. By varying the amount of labeled samples, we get insights into how performance decreases in very limited label regimes.

For evaluation, we randomly split the available data in $\mathcal{D}_{train}$ and $\mathcal{D}_{test}$ using a class stratified split into 90\% training and 10\% test data, ensuring that the class distribution is preserved across both subsets. We train each model $f(\cdot|\theta)$ on $\mathcal{D}_{train}$ with varying amounts of labeled samples $N_l\in \{2595, 3893, 5190, 12973, 25946\}.$ Model performance is compared on $\mathcal{D}_{test}$. To exclude the influence of unfavorably selected labeled samples, we repeat the unlabeling step $\mathcal{D}=(\mathcal{D}_{l},\mathcal{D}_{ul})$ five times, which is also in line with \cite{goschenhofer2022deep}.
\subsection{Performance Metrics}\label{sec:metrics}
Performance evaluation is particularly important for highly imbalanced datasets such as those encountered in fraud detection \cite{dalpoz2017credit}. In this setting, only a very small fraction of transactions can be manually invested or automatically blocked, meaning that model performance in the head of the ranked distribution is of primary interest. Recent work by \cite{hayat2025data} has shown that the choice of evaluation methodology often has a greater impact on reported performance than model complexity itself. 

In this work, we adopt a comprehensive set of evaluation metrics combining standard global measures for imbalanced data with domain-specific, cost-sensitive metrics. The global measures include area under the precision recall curve (PR-AUC), macro F1 score and cross entropy loss. Although providing an initial overview of overall performance, such global metrics are less informative in practice because they aggregate over thresholds that are rarely relevant for real-world operation \cite{dalpoz2017credit}. To address these limitations, and in line with common practice in information retrieval \cite{manning2008introduction} and fraud detection research \cite{bahnsen2016feature, dalpoz2017credit}, we complement them with the head metrics Precision@K, Recall@K, partial PR-AUC, and Expected Cost@K to directly reflect operational constraints.
Let TP, FP, TN, FN denote the number of true positives, false positives, true negatives and false negatives, respectively. Then precision and recall are defined as $\text{Precision} = \mathrm{TP}/(\mathrm{TP}+\mathrm{FP})$ and $\text{Recall} = \mathrm{TP}/(\mathrm{TP}+\mathrm{FN})$. The F1 score combined them as
\begin{align*}
\text {F1}&=2 \frac{\text {Precision} \cdot \text {Recall}}{\text {Precision}+ \text {Recall}}=\frac{2 \mathrm{TP}}{2 \mathrm{TP}+\mathrm{FP}+\mathrm{FN}}.
\end{align*}
The macro F1 score is obtained by averaging the F1 scores for the two classes (fraud and non-fraud), ensuring equal importance of both despite the class imbalance. The PR curve plots precision as a function of recall and PR-AUC, denoting the area under this curve, summarizing performance across thresholds. 

Beyond point classification, we also assess the quality of predictive distributions using the cross entropy loss. For $N$ transactions, it is given by
\[\text{Cross Entropy} = -\frac{1}{N} \sum_{i=1}^N \Big( \mathds{1}_{\left\{a_i=fraud\right\}} \cdot \log(f_i) + (1-\mathds{1}_{\left\{a_i=fraud\right\}} ) \cdot \log(1-f_i) \Big),\]
where $f_{i}$ denotes the predicted probability that transaction $a_i$ is fraudulent.

To focus on the operationally most relevant part of the distribution, we define head metrics with respect to the top $K\%$ of transactions ranked by predicted fraud probability. Particularly,
\begin{align*}
\text{Precision@K} &= \frac{\mathrm{TP}_{@K}}{\mathrm{TP}_{@K} + \mathrm{FP}_{@K}}, \\
\text{Recall@K}    &= \frac{\mathrm{TP}_{@K}}{\mathrm{TP}_{@K} + \mathrm{FN}_{@K}},
\end{align*}	
where $\mathrm{TP}_{@K}$ and $\mathrm{FP}_{@K}$ denote the number of true and false positives within the top $K\%$ and $\mathrm{FN}_{@K}$ denotes the false negatives outside of this subset. 
We additionally employ partial PR-AUC, a restricted variant of standard PR-AUC. Instead of integrating precision over the full recall range $[0,1]$, partial PR-AUC is computed only over a recall interval $[0,r]$ for fixed $r<1$ as,
\[\text{Partial PR-AUC}_{r} = \int_0^{r} P(s)\,ds,\]
where $P(s)$ denotes the precision at recall level $s$. This restriction reflects the reality that financial institutions rarely operate at very high recall values due to limited investigation capacity. By focusing on realistic recall ranges, partial PR-AUC better reflects the ranking quality of models in the operational regime. 

Finally, to incorporate financial impact, we adopt Expected Cost@K in line with cost-sensitive learning frameworks \cite{elkan2001foundations, bahnsen2016feature}. 
A false negative (missed fraud) is assigned the full amount, representing reimbursement to the customer, whereas a false positive (legitimate transaction flagged as fraud) is assigned a fixed fraction $\alpha=0.02$ of the transaction, representing the lost transaction fees and operational overhead. For transaction amounts $a_i$ this yields,
\[\text{Cost@K} = \sum_{i \in \text{FN}_{@K}} a_i \;+\; \alpha \sum_{i \in \text{FP}_{@K}} a_i.\]
This cost-based measure enables direct comparison of models in terms of their business impact, completing the global statistical performance metrics.

\subsection{Runtime and Scalability}
All experiments were conducted on Kaggle's free Tesla T4 GPU, without access to high-performance clusters. On the BankSim dataset with about 600k transactions end-to-end training for an individual model requires approximately 20-30 minuts depending on the amount of labeled samples. Despite the Bayesian framework, training remains efficient due the shallow architecture of both generator and discriminator, highlighting practical deployability. Codes for reproducing the results and figures for this paper are available on \url{https://github.com/DavidHirnschall/logsig-bayesian-gan}.

\subsection{Numerical Results}
\subsubsection{Discriminative performance evaluation}
We evaluate the performance of all trained models using labeled subsets of size $N_l\in \{2595, 3893, 5190, 12973, 25946\}$, corresponding to 0.5$\%$, 0.75$\%$, 1$\%$, 2.5$\%$ and 5$\%$ of available training samples. The labeled subsets were randomly selected while preserving the original class distribution, resulting in $N_f=29, 44, 58, 144, 288$ fraudulent transactions, respectively. 
To obtain a global comparison across different training sizes, we report three complementary metrics, namely Macro F1, PR-AUC and cross-entropy loss (CEL) (see \cref{sec:metrics}). These capture balanced classification performance across classes, ranking quality under class imbalance and probabilistic calibration. As expected, we observe higher volatility across all metrics and models for smaller labeled subsets, with performance stabilizing as $N_l$ increases. Further, the fully supervised neural network (FNN) displays a steeper performance curve, gradually catching up, and in some cases surpassing our semi-supervised model as the amount of labeled data grows and the benefit of semi-supervised learning diminishes.\\
Macro F1 results are reported in \cref{tab:macro_f1}. Across all sample sizes either, our proposed approach or the FNN achieved the highest scores.
\begin{table}[H]
	\centering
	\caption{Macro F1 scores across models and training sizes. Values are mean ($\pm$ std). Best values are in bold.}
	\label{tab:macro_f1}
	\scalebox{0.7}{
		\begin{tabular}{c|cccccccc}
			\toprule
			\textbf{$N_l$} & \textbf{NB} & \textbf{LReg} & \textbf{LReg-SSL} & \textbf{KNN} & \textbf{SVM} & \textbf{RF} & \textbf{FN} & \textbf{Our model} \\
			\midrule
			2595  & 0.640 ($\pm$0.020) & 0.532 ($\pm$0.024) & 0.505 ($\pm$0.003) & 0.536 ($\pm$0.026) & 0.640 ($\pm$0.032) & 0.691 ($\pm$0.054) & 0.808 ($\pm$0.024) & \textbf{0.810 ($\pm$0.024)} \\
			3893  & 0.639 ($\pm$0.015) & 0.530 ($\pm$0.021) & 0.510 ($\pm$0.006) & 0.548 ($\pm$0.026) & 0.692 ($\pm$0.096) & 0.725 ($\pm$0.039) & \textbf{0.819 ($\pm$0.017)} & 0.819 ($\pm$0.021) \\
			5190  & 0.641 ($\pm$0.008) & 0.531 ($\pm$0.007) & 0.515 ($\pm$0.007) & 0.548 ($\pm$0.018) & 0.728 ($\pm$0.043) & 0.742 ($\pm$0.040) & 0.830 ($\pm$0.012) & \textbf{0.833 ($\pm$0.014)} \\
			12973 & 0.643 ($\pm$0.004) & 0.580 ($\pm$0.011) & 0.548 ($\pm$0.004) & 0.600 ($\pm$0.006) & 0.800 ($\pm$0.027) & 0.817 ($\pm$0.011) & 0.855 ($\pm$0.008) & \textbf{0.856 ($\pm$0.004)} \\
			25946 & 0.645 ($\pm$0.004) & 0.617 ($\pm$0.006) & 0.560 ($\pm$0.003) & 0.644 ($\pm$0.017) & 0.822 ($\pm$0.019) & 0.837 ($\pm$0.007) & \textbf{0.864 ($\pm$0.005)} & 0.858 ($\pm$0.004) \\
			\bottomrule
		\end{tabular}
	}
\end{table}
In terms of CEL, our model consistently outperformed all benchmark models across all labeled training sizes (\cref{tab:cel}). However, we again observe a narrowing gap as the FNN benefits from larger datasets.
\begin{table}[H]
	\centering
	\caption{Cross Entropy loss across models and training sizes. Values are mean ($\pm$ std.). Best values per row are in bold.}
	\label{tab:cel}
	\scalebox{0.7}{
		\begin{tabular}{c|cccccccc}
			\toprule
			\textbf{$N_l$} & \textbf{NB} & \textbf{LReg} & \textbf{LReg-SSL} & \textbf{KNN} & \textbf{SVM} & \textbf{RF} & \textbf{FN} & \textbf{Our model} \\
			\midrule
			2595  & 1.332 ($\pm$0.226) & 0.048 ($\pm$0.001) & 0.094 ($\pm$0.002) & 0.333 ($\pm$0.018) & 0.048 ($\pm$0.013) & 0.046 ($\pm$0.005) & 0.051 ($\pm$0.004) & \textbf{0.032 ($\pm$0.002)} \\
			3893  & 1.371 ($\pm$0.193) & 0.047 ($\pm$0.000) & 0.087 ($\pm$0.001) & 0.316 ($\pm$0.012) & 0.042 ($\pm$0.008) & 0.044 ($\pm$0.003) & 0.043 ($\pm$0.002) & \textbf{0.028 ($\pm$0.002)} \\
			5190  & 1.334 ($\pm$0.091) & 0.045 ($\pm$0.000) & 0.082 ($\pm$0.001) & 0.301 ($\pm$0.008) & 0.035 ($\pm$0.003) & 0.044 ($\pm$0.001) & 0.036 ($\pm$0.005) & \textbf{0.025 ($\pm$0.001)} \\
			12973 & 1.313 ($\pm$0.041) & 0.042 ($\pm$0.000) & 0.067 ($\pm$0.000) & 0.265 ($\pm$0.011) & 0.031 ($\pm$0.004) & 0.040 ($\pm$0.002) & 0.024 ($\pm$0.002) & \textbf{0.021 ($\pm$0.001)} \\
			25946 & 1.295 ($\pm$0.059) & 0.038 ($\pm$0.000) & 0.057 ($\pm$0.001) & 0.242 ($\pm$0.019) & 0.029 ($\pm$0.003) & 0.038 ($\pm$0.003) & \textbf{0.020 ($\pm$0.001)} & \textbf{0.020 ($\pm$0.000)} \\
			\bottomrule
		\end{tabular}
	}
\end{table}
For PR-AUC, the random forest achieved the highest overall scores, followed closely by our model and the FNN (see \cref{tab:prauc}).

\begin{table}[H]
	\centering
	\caption{PR-AUC across models and training sizes. Values are mean ($\pm$ std.). Best values per row are in bold.}
	\label{tab:prauc}
	\scalebox{0.7}{
		\begin{tabular}{c|cccccccc}
			\toprule
			\textbf{$N_l$} & \textbf{NB} & \textbf{LReg} & \textbf{LReg-SSL} & \textbf{KNN} & \textbf{SVM} & \textbf{RF} & \textbf{FN} & \textbf{Our model} \\
			\midrule
			2595  & 0.498 ($\pm$0.004) & 0.155 ($\pm$0.011) & 0.194 ($\pm$0.024) & 0.163 ($\pm$0.023) & 0.417 ($\pm$0.078) & \textbf{0.661 ($\pm$0.060)} & 0.588 ($\pm$0.059) & 0.620 ($\pm$0.042) \\
			3893  & 0.503 ($\pm$0.006) & 0.187 ($\pm$0.017) & 0.236 ($\pm$0.026) & 0.214 ($\pm$0.020) & 0.500 ($\pm$0.102) & \textbf{0.683 ($\pm$0.041)} & 0.628 ($\pm$0.040) & 0.658 ($\pm$0.032) \\
			5190  & 0.505 ($\pm$0.004) & 0.209 ($\pm$0.018) & 0.254 ($\pm$0.015) & 0.239 ($\pm$0.023) & 0.580 ($\pm$0.031) & \textbf{0.696 ($\pm$0.039)} & 0.663 ($\pm$0.026) & 0.693 ($\pm$0.019) \\
			12973 & 0.505 ($\pm$0.003) & 0.300 ($\pm$0.018) & 0.338 ($\pm$0.013) & 0.359 ($\pm$0.021) & 0.663 ($\pm$0.040) & \textbf{0.754 ($\pm$0.012)} & 0.742 ($\pm$0.014) & 0.752 ($\pm$0.009) \\
			25946 & 0.504 ($\pm$0.003) & 0.384 ($\pm$0.015) & 0.407 ($\pm$0.013) & 0.443 ($\pm$0.037) & 0.688 ($\pm$0.022) & \textbf{0.772 ($\pm$0.010)} & 0.772 ($\pm$0.009) & 0.767 ($\pm$0.005) \\
			\bottomrule
		\end{tabular}
	}
\end{table}

\begin{figure}[ht]
	\centering
	\includegraphics[width=0.8\linewidth]{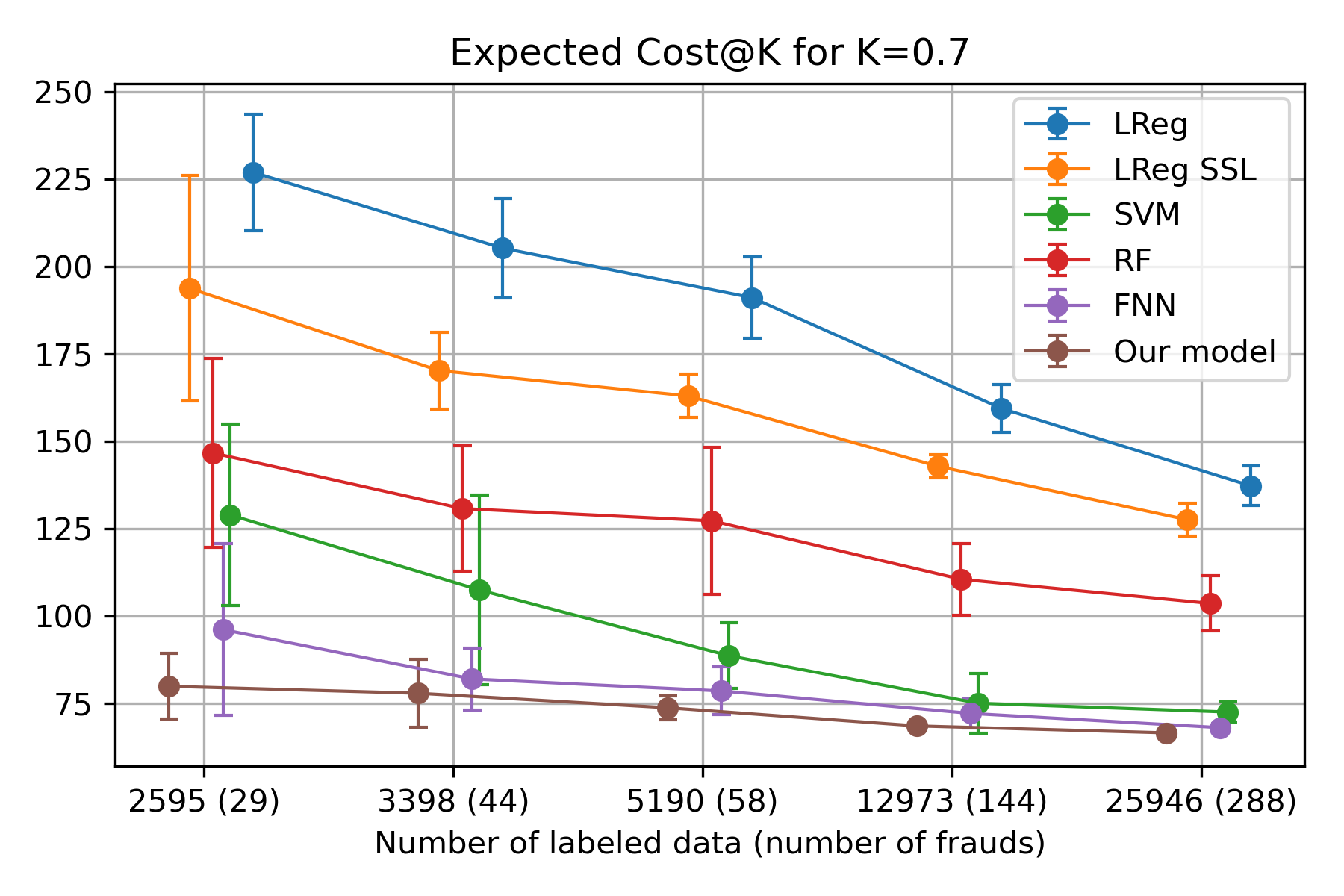}
	\caption{Expected Cost@K for K=0.5 for various amounts of labeled samples $N_l\in \{2595, 3893, 5190, 12973, 25946\}.$ Dots represent mean Expected Cost@Ks and the vertical error bars are the standard deviation of 5 repeated random unlabelings. Dots are jittered on the x-axis to avoid overlapping error bars. Values are presented in thousands.}
	\label{fig:cost} 
\end{figure}

While the given global metrics provide a useful statistical model comparison, in financial fraud detection, the expected financial cost of errors is often more relevant in practice. Therefore, we, adopt Expected Cost@K as our primary domain-specific metric, where $K$ controls the fraction of transactions to be blocked or investigated. A visual comparison for $K=0.5\%$ is given in \cref{fig:cost}. Our model achieved the lowest expected cost, reducing financial cost relative to the statistically strong RF baseline by approximately 35-45\% across labeled training sizes. A detailed comparison for multiple values of $K$ is reported in \cref{tab:appendix_costK}.

We further report Precision@K for $K=0.5$ in \cref{tab:precisionk} Our model achieved higher scores for limited labeled sample sizes, whereas the FNN gradually overtakes as more labeled samples become available. Full results for additional $k$ values are presented in \cref{tab:appendix_precisionK}.
\begin{table}[ht]
	\centering
	\caption{Precision@K for $K=0.5$ for different models and training sizes. Values are mean ($\pm$ std.). Bold numbers indicate the best performance per row.}
	\label{tab:precisionk}
	\scalebox{0.7}{
		\begin{tabular}{c|ccccccccc}
			\toprule
			\textbf{$N_l$} & \textbf{NB} & \textbf{L} & \textbf{L-SSL} & \textbf{KNN} & \textbf{SVM} & \textbf{RF} & \textbf{FN} & \textbf{Our model} \\
			\midrule
			2595  & 0.201 ($\pm$0.027) & 0.331 ($\pm$0.036) & 0.394 ($\pm$0.045) & 0.266 ($\pm$0.033) & 0.594 ($\pm$0.123) & 0.819 ($\pm$0.079) & 0.833 ($\pm$0.055) & \textbf{0.883 ($\pm$0.044)} \\
			3893  & 0.216 ($\pm$0.075) & 0.366 ($\pm$0.027) & 0.441 ($\pm$0.029) & 0.289 ($\pm$0.056) & 0.709 ($\pm$0.138) & 0.852 ($\pm$0.051) & 0.877 ($\pm$0.038) & \textbf{0.896 ($\pm$0.044)} \\
			5190  & 0.225 ($\pm$0.042) & 0.392 ($\pm$0.030) & 0.469 ($\pm$0.010) & 0.305 ($\pm$0.073) & 0.802 ($\pm$0.046) & 0.880 ($\pm$0.036) & 0.910 ($\pm$0.014) & \textbf{0.919 ($\pm$0.020)} \\
			12973 & 0.216 ($\pm$0.018) & 0.490 ($\pm$0.032) & 0.541 ($\pm$0.017) & 0.442 ($\pm$0.035) & 0.889 ($\pm$0.058) & 0.920 ($\pm$0.013) & \textbf{0.945 ($\pm$0.008)} & 0.944 ($\pm$0.005) \\
			25946 & 0.188 ($\pm$0.031) & 0.583 ($\pm$0.034) & 0.621 ($\pm$0.024) & 0.549 ($\pm$0.048) & 0.898 ($\pm$0.024) & 0.938 ($\pm$0.012) & \textbf{0.958 ($\pm$0.009)} & 0.952 ($\pm$0.008) \\
			\bottomrule
		\end{tabular}
	}
\end{table}

Similarly, Recall@K for $K=0.5$ is reported in \cref{tab:recallk}. Our model performs best for limited labeled sample sizes, while the FNN surpasses it once larger sample sizes are used. Results for additional values for Additional Recall@K results are given in \cref{tab:appendix_recallK}.

\begin{table}[ht]
	\centering
	\caption{Recall@K for $K=0.5$ for different models and training sizes. Values are mean ($\pm$ std.). Bold numbers indicate the best performance per row.}
	\label{tab:recallk}
	\scalebox{0.7}{
		\begin{tabular}{c|cccccccc}
			\toprule
			\textbf{$N_l$} & \textbf{NB} & \textbf{L} & \textbf{L-SSL} & \textbf{KNN} & \textbf{SVM} & \textbf{RF} & \textbf{FN} & \textbf{Our model} \\
			\midrule
			2595  & 0.091 ($\pm$0.012) & 0.150 ($\pm$0.016) & 0.178 ($\pm$0.020) & 0.120 ($\pm$0.015) & 0.269 ($\pm$0.056) & 0.370 ($\pm$0.036) & 0.377 ($\pm$0.025) & \textbf{0.399 ($\pm$0.020)} \\
			3893  & 0.098 ($\pm$0.034) & 0.166 ($\pm$0.012) & 0.199 ($\pm$0.013) & 0.131 ($\pm$0.025) & 0.321 ($\pm$0.062) & 0.385 ($\pm$0.023) & 0.397 ($\pm$0.017) & \textbf{0.405 ($\pm$0.020)} \\
			5190  & 0.102 ($\pm$0.019) & 0.177 ($\pm$0.014) & 0.212 ($\pm$0.005) & 0.138 ($\pm$0.033) & 0.363 ($\pm$0.021) & 0.398 ($\pm$0.016) & 0.412 ($\pm$0.006) & \textbf{0.416 ($\pm$0.009)} \\
			12973 & 0.098 ($\pm$0.008) & 0.222 ($\pm$0.014) & 0.245 ($\pm$0.008) & 0.200 ($\pm$0.016) & 0.402 ($\pm$0.026) & 0.416 ($\pm$0.006) & \textbf{0.428 ($\pm$0.003)} & 0.427 ($\pm$0.002) \\
			25946 & 0.085 ($\pm$0.014) & 0.264 ($\pm$0.015) & 0.281 ($\pm$0.011) & 0.248 ($\pm$0.022) & 0.406 ($\pm$0.011) & 0.424 ($\pm$0.005) & \textbf{0.433 ($\pm$0.004)} & 0.431 ($\pm$0.004) \\
			\bottomrule
		\end{tabular}
	}
\end{table}

Finally, partial PR-AUC with a recall threshold $r=0.7$ is reported in \cref{tab:prauck}. Here, our model achieved the best performance across all sample sizes except the largest, where the FNN wins.
Extended results for different recall thresholds are provided in \cref{tab:appendix_prauck}.

\begin{table}[H]
	\centering
	\caption{Partial PR-AUC at $r=0.7$ across different training sizes.Values are mean ($\pm$ std.). Bold numbers indicate the best performance per row.}
	\label{tab:prauck}
	\scalebox{0.7}{
		\begin{tabular}{c|cccccccc}
			\toprule
			\textbf{$N_l$} & \textbf{NB} & \textbf{L} & \textbf{L-SSL} & \textbf{KNN} & \textbf{SVM} & \textbf{RF} & \textbf{FN} & \textbf{Our model} \\
			\midrule
			2595  & 0.458 ($\pm$0.001) & 0.147 ($\pm$0.013) & 0.183 ($\pm$0.025) & 0.375 ($\pm$0.026) & 0.395 ($\pm$0.077) & 0.569 ($\pm$0.051) & 0.560 ($\pm$0.043) & \textbf{0.582 ($\pm$0.038)} \\
			3893  & 0.459 ($\pm$0.001) & 0.176 ($\pm$0.016) & 0.224 ($\pm$0.027) & 0.399 ($\pm$0.012) & 0.471 ($\pm$0.099) & 0.590 ($\pm$0.032) & 0.590 ($\pm$0.027) & \textbf{0.606 ($\pm$0.027)} \\
			5190  & 0.459 ($\pm$0.001) & 0.197 ($\pm$0.017) & 0.240 ($\pm$0.015) & 0.413 ($\pm$0.018) & 0.538 ($\pm$0.028) & 0.602 ($\pm$0.028) & 0.616 ($\pm$0.013) & \textbf{0.624 ($\pm$0.013)} \\
			12973 & 0.459 ($\pm$0.001) & 0.283 ($\pm$0.017) & 0.318 ($\pm$0.013) & 0.472 ($\pm$0.014) & 0.603 ($\pm$0.038) & 0.637 ($\pm$0.007) & 0.649 ($\pm$0.007) & \textbf{0.651 ($\pm$0.006)} \\
			25946 & 0.459 ($\pm$0.000) & 0.363 ($\pm$0.014) & 0.383 ($\pm$0.012) & 0.513 ($\pm$0.016) & 0.618 ($\pm$0.021) & 0.649 ($\pm$0.007) & \textbf{0.660 ($\pm$0.003)} & 0.657 ($\pm$0.004) \\
			\bottomrule
		\end{tabular}
	}
\end{table}

While head metrics emphasize the practically relevant head of the ranked distribution, relying solely on point estimates may still be misleading. A single prediction may appear highly confident, placing a transaction high in the ranking, yet posterior sampling may reveal substantial uncertainty of the underlying model and correct such spurious point estimates and push uncertain cases lower in the ranking. This motivates the following analysis of Bayesian uncertainty, where predictive distributions rather than single predictions provide a more robust basis for decision-making.

\subsubsection{Uncertainty evaluation}
To assess predictive uncertainty of our Bayesian approach, we approximate the predictive distribution for each transaction over fraud probability $f_i$ from SGHMC samples and summarize uncertainty by the 90\% posterior interval width $u_i=Q_{0.95}-Q_{0.05}.$ In practice, one may call a prediction uncertain if the classification threshold $\tau$ lies within the interval $[Q_{0.05}, Q_{0.95}]$, as it returns mixed classification signals. In the quantitative evaluation below, we use the continuous score $u_i.$

\begin{figure}[ht]
	\centering
	\includegraphics[width=0.6\textwidth]{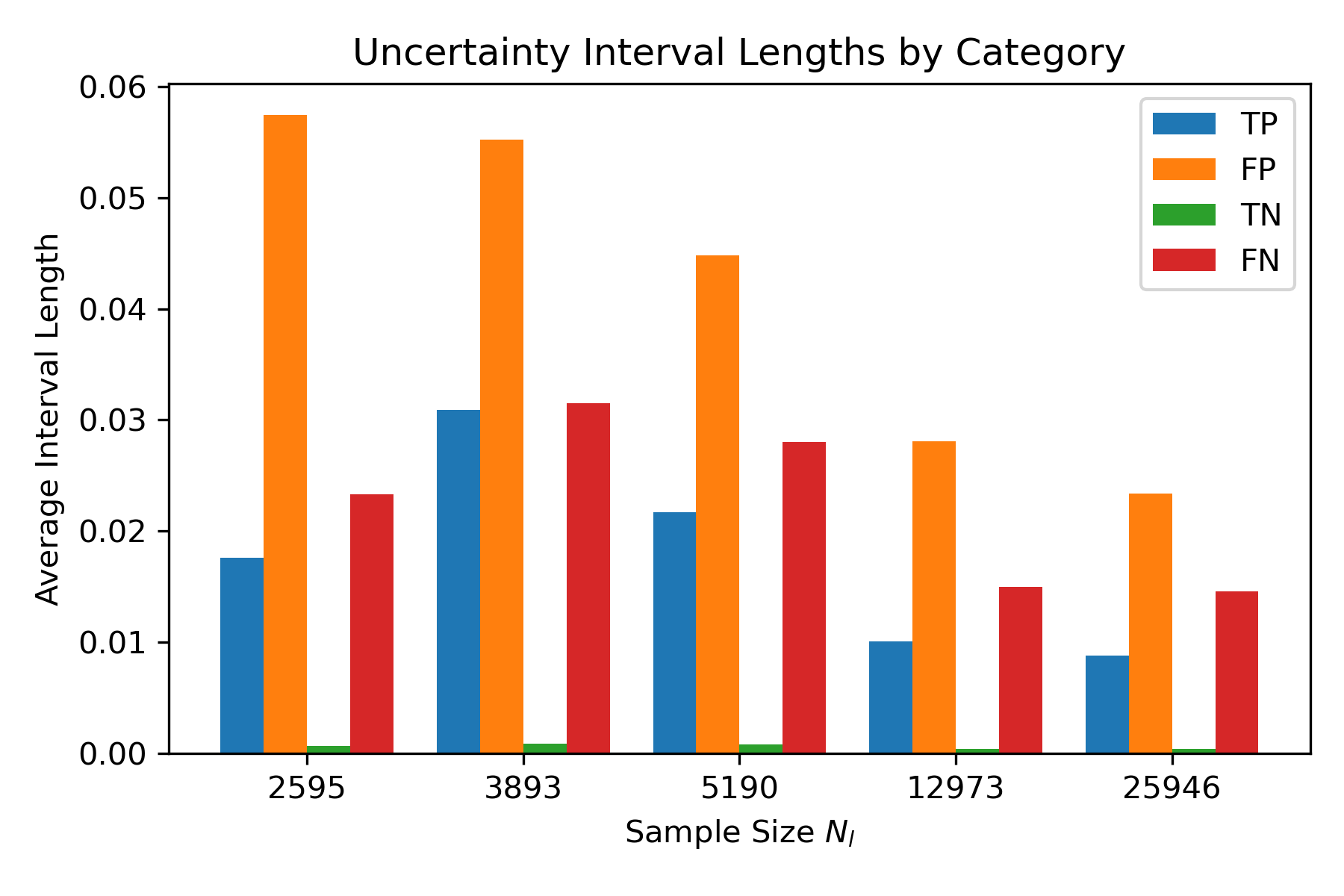}
	\caption{Average 90\% uncertainty interval width over five unlabelings by outcome (TP/FP/TN/FN) across labeled sample sizes $N_l$.  Misclassified instances (FP/FN) are consistently more uncertain then correct ones (TP/TN).}
	\label{fig:widths}
\end{figure}

To test whether uncertainty identifies mistakes in predictions, we treat misclassifications as positive class, compute the ROC curve using the uncertainty scores $u_i$ as the ranking variable and report the area under the ROC curve (AUROC) in \cref{tab:auroc}. While the ROC curve plots the true positive rate against the false positive rate across thresholds, the AUROC gives the probability that an error receives a higher uncertainty score than a correct prediction by
\[\mathrm{AUROC} = \Pr\!\big(u_{\text{error}} > u_{\text{correct}}\big)+\tfrac12\,\Pr\!\big(u_{\text{error}} = u_{\text{correct}}\big)w\]
with 0.5 indicating random ranking and higher scores better performance.
\begin{table}[H]
	\centering
	\caption{AUROC values across sample sizes $N_l$.}
	\label{tab:auroc}
	\begin{tabular}{c|c|c|c|c|c}
		\toprule
		$N_l$ & 2595 & 3893 & 5190 & 12973 & 25946 \\
		\midrule
		AUROC & 0.8730 & 0.9010 & 0.9132 & 0.9307 & 0.9345 \\
		\bottomrule
	\end{tabular}
\end{table}
Further, we categorize transactions in true positive (TP), false positive (FP), true negative (TN) and false negative (FN) via the posterior mean prediction $\bar f_i$. \cref{fig:widths} displays average 90\% interval widths over five unlabelings for each category (TP, FP, TN, FN) and across labeled sample sizes. Misclassified instances (FP and FN) consistently exhibit a substantially larger uncertainty than correctly classified ones (TP and TN), supporting risk-aware decision policies in high-risk settings.

\begin{figure}[ht]
	\centering
	\subfloat[Two samples classified as fraud]{\includegraphics[width=.5\textwidth]{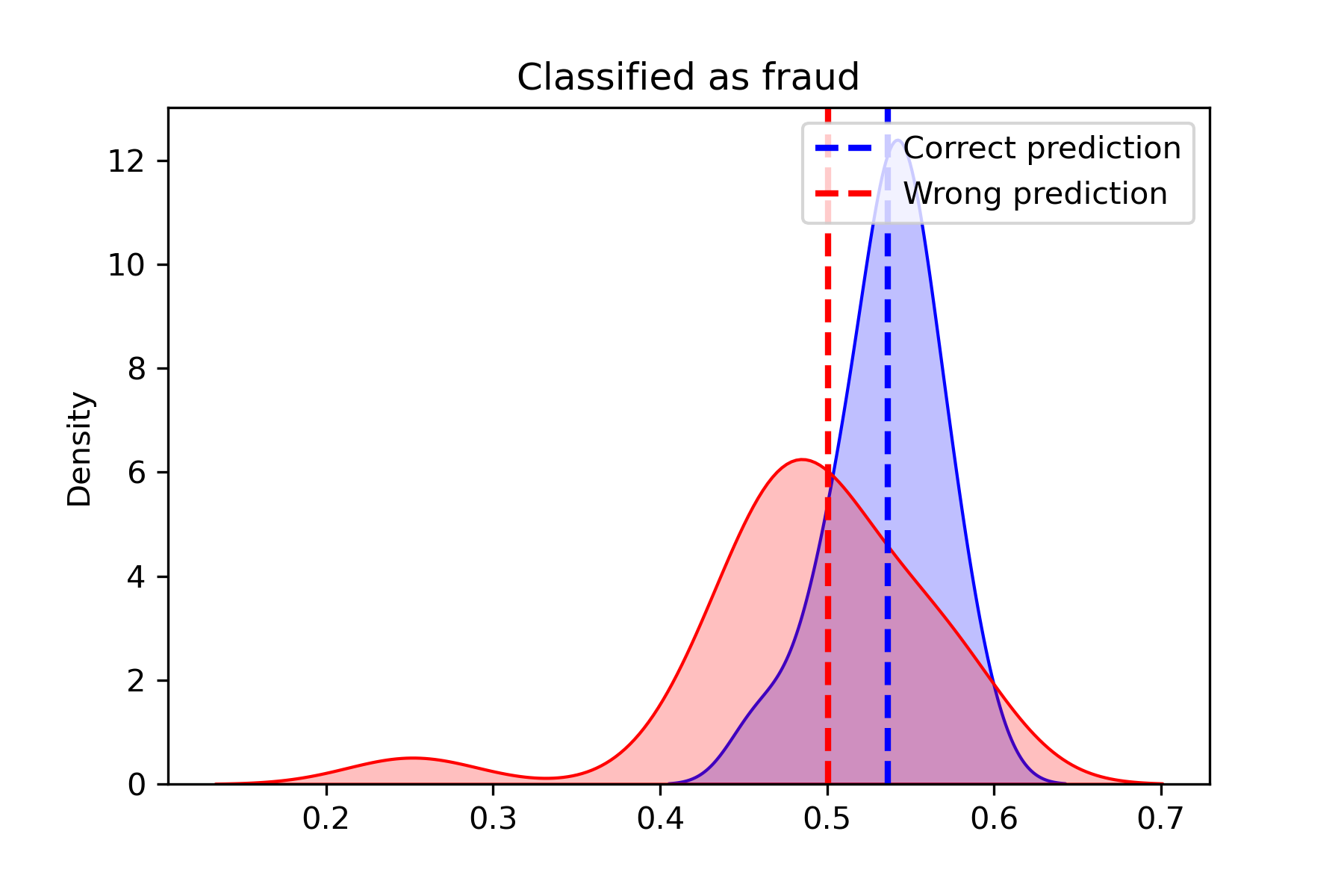}}	
	\subfloat[Two samples classified as non-fraud]{\includegraphics[width=.5\textwidth]{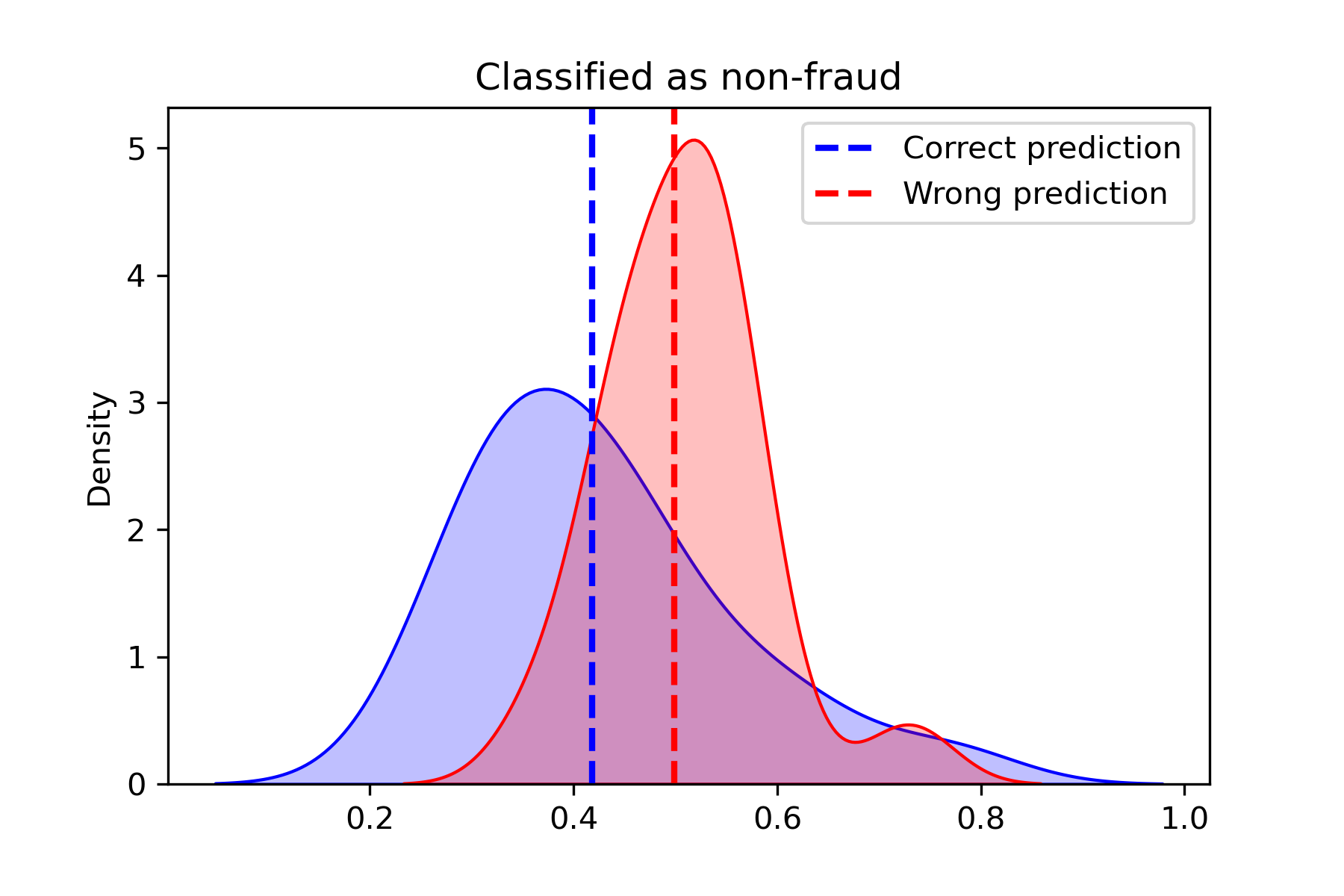}}
	\caption{Predictive distributions for four uncertain predictions, one from each category (TP, FP, TN, FN).}
	\label{uncertainty_individual}
\end{figure}

Finaly, \cref{uncertainty_individual} shows posterior distributions for four representative transactions (one per outcome category) with decision threshold $\tau$ overlaid by the uncertainty interval.  Among samples labeled as fraud, misclassified FPs tend to show broader posteriors, hence higher uncertainty, with more weight in the left tail. Among transactions labeled as non-fraud, misclassified FNs exhibit wider predictive posteriors with heavier right tails. These pattern reflect epistemic uncertainty beyond the point estimate $\bar f_i$.

\section{Conclusion and Discussion}\label{conclusion}
In this paper, we introduced a novel deep generative semi-supervised approach for time series classification that leverages conditional GANs, Bayesian inference, and log-signatures to address core challenges in financial fraud detection: irregularly sampled data of varying length, limited labeled samples and the need for probabilistic predictions with uncertainty quantification. Log-signatures provide a principled way to encode transaction histories of variable length, enabling robust learning where other sequence models struggle.

To provide a comprehensive performance assessment, we combined global statistical metrics (Macro F1, PR-AUC, Cross-Entropy loss) with domain-specific head metrics (Precision@K, Recall@K, partial PR-AUC, Expected Cost@K). This dual evaluation framework reflects both statistical overall performance and real-world business impact, where only a small fraction of transactions can be reviewed. Our empirical evaluation on the BankSim dataset demonstrated that our proposed approach outperforms established baselines in the low-data regime, with particularly strong gains in cost-based performance, achieving up to 45\% lower Expected Cost@K than strong statistical performers such as random forest. While fully supervised neural networks close the performance gap as more labeled samples become available, our approach maintains a clear advantage when labeled data is scarce.

Another key contribution lies in uncertainty quantification. By placing a distribution over network weights, our Bayesian framework produces predictive distributions rather than point estimates, allowing calibrated confidence intervals. Misclassified samples were shown to exhibit consistently higher uncertainty, highlighting the importance of uncertainty-aware predictions in high-risk domains where wrong decisions carry substantial cost. 

Nonetheless, several limitations remain. Our approach relies on transaction histories, making predictions for new or low-activity customers challenging. Moreover, fraud detection assumes that fraud breaks behavioral patterns, which may fail in cases of repeated fraud that paradoxically resemble a customer’s transaction history. Finally, our evaluation relies on the BankSim dataset. While widely used in fraud detection for its realism, it remains synthetic, and future work should extend validation to real-world datasets as access becomes possible. As with most data-driven systems, model generalization depends on representative training data and evolving fraud tactics will require continuous retraining.

Overall, our findings demonstrate that semi-supervised Bayesian generative models, combined with log-signatures for temporal feature encoding, can effectively handle variable-length sequences of irregular sampling frequencies while providing robust, uncertainty-aware decision support. This synergy offers tangible benefits for fraud detection and broader time series classification tasks.

\appendix
\section{Full Experimental Results}
\cref{tab:scaling_effects} shows the effect of scaling the transaction amount feature on classical baselines. Across all sizes of labeled training sets, performance drops dramatically when scaling is applied. Naive Bayes collapses to random-like performance with a macro F1 score of about 0.1, logistic regression suffers orders of magnitude worse calibration (cross entropy loss increased from under 0.1 to over 5) and PR-AUC values shrink significantly. This highlights the sensitivity to raw values, indicating that naive preprocessing can destroy predictive signal.

\begin{table}[H]
	\centering
	\caption{Performance of baseline models with and without scaling of the amount feature. Values are means across random unlabelings.}
	\label{tab:scaling_effects}
	\scalebox{0.7}{
		\begin{tabular}{c|c|cc|cc|cc|cc|cc|cc}
			\toprule
			\multirow{2}{*}{$N_l$} & \multirow{2}{*}{Metric} 
			& \multicolumn{2}{c|}{NB} 
			& \multicolumn{2}{c|}{LReg} 
			& \multicolumn{2}{c|}{LReg-SSL} 
			& \multicolumn{2}{c|}{KNN} 
			& \multicolumn{2}{c|}{SVM} 
			& \multicolumn{2}{c}{RF} \\ 
			\cline{3-14}
			& & Unscaled & Scaled & Unscaled & Scaled & Unscaled & Scaled & Unscaled & Scaled & Unscaled & Scaled & Unscaled & Scaled \\ 
			\midrule
			\multirow{3}{*}{2595} 
			& Macro F1 & 0.640 & 0.013 & 0.532 & 0.215 & 0.505 & 0.376 & 0.536 & 0.499 & 0.640 & 0.497 & 0.691 & 0.558 \\
			& CEL      & 1.332 & 33.987 & 0.048 & 8.163 & 0.094 & 5.105 & 0.333 & 0.373 & 0.048 & 0.062 & 0.046 & 0.488 \\
			& PRAUC    & 0.498 & 0.506 & 0.155 & 0.558 & 0.194 & 0.577 & 0.163 & 0.035 & 0.417 & 0.082 & 0.661 & 0.277 \\ \hline
			
			\multirow{3}{*}{3893} 
			& Macro F1 & 0.639 & 0.013 & 0.530 & 0.200 & 0.510 & 0.314 & 0.548 & 0.502 & 0.692 & 0.508 & 0.725 & 0.612 \\
			& CEL      & 1.371 & 33.992 & 0.047 & 8.509 & 0.087 & 6.926 & 0.316 & 0.365 & 0.042 & 0.062 & 0.044 & 0.493 \\
			& PRAUC    & 0.503 & 0.506 & 0.187 & 0.548 & 0.236 & 0.548 & 0.214 & 0.056 & 0.500 & 0.166 & 0.683 & 0.355 \\ \hline
			
			\multirow{3}{*}{5190} 
			& Macro F1 & 0.641 & 0.013 & 0.531 & 0.366 & 0.515 & 0.329 & 0.548 & 0.502 & 0.728 & 0.515 & 0.742 & 0.530 \\
			& CEL      & 1.334 & 33.992 & 0.045 & 5.117 & 0.082 & 6.125 & 0.301 & 0.361 & 0.035 & 0.061 & 0.044 & 0.507 \\
			& PRAUC    & 0.505 & 0.506 & 0.209 & 0.610 & 0.254 & 0.557 & 0.239 & 0.062 & 0.580 & 0.229 & 0.696 & 0.369 \\ \hline
			
			\multirow{3}{*}{12973} 
			& Macro F1 & 0.643 & 0.013 & 0.580 & 0.152 & 0.548 & 0.338 & 0.600 & 0.511 & 0.800 & 0.520 & 0.817 & 0.471 \\
			& CEL      & 1.313 & 33.998 & 0.042 & 14.521 & 0.067 & 4.416 & 0.265 & 0.344 & 0.031 & 0.060 & 0.040 & 0.596 \\
			& PRAUC    & 0.505 & 0.506 & 0.300 & 0.528 & 0.338 & 0.573 & 0.359 & 0.100 & 0.663 & 0.212 & 0.754 & 0.440 \\ \hline
			
			\multirow{3}{*}{25946} 
			& Macro F1 & 0.645 & 0.013 & 0.617 & 0.229 & 0.560 & 0.225 & 0.644 & 0.525 & 0.822 & 0.522 & 0.837 & 0.597 \\
			& CEL      & 1.295 & 33.997 & 0.038 & 6.422 & 0.057 & 11.258 & 0.242 & 0.329 & 0.029 & 0.060 & 0.038 & 0.546 \\
			& PRAUC    & 0.504 & 0.506 & 0.384 & 0.568 & 0.407 & 0.524 & 0.443 & 0.146 & 0.688 & 0.191 & 0.772 & 0.518 \\
			\bottomrule
	\end{tabular}}
\end{table}

Following \cite{heusel2017gans}, who highlights the ne need for a higher learning rate in the discriminator, as it must adapt rapidly to both classification and discrimination, while the generator only learns indirectly through the discriminators feedback, we fixed the generators learning rate to a standard $10^{-4}$ and tuned the discriminators learning rate as a multiple thereof. We performed a small calibration study using three multipliers (1, 10, 50) across three labeled sample sizes and only three random unlabelings. For each configuration we report mean $\pm$ standard deviation for Expected Cost@K with $K=0.5$, macro F1 and cross-entropy loss. \cref{tab:lr_tuning} show that $lr=0.005$ achieved comparable mean performance to $lr=0.001$, with only a modest increase in variability. As higher learning rates typically accelerate convergence, and since our main results average over five random unlabelings, hence provinde more stability, we opt for the higher learning rate of 0.005. 

\begin{table}[H]
	\centering
	\caption{Performance of our model under different learning rates. Values are reported as mean $\pm$ std across random unlabelings.}
	\label{tab:lr_tuning}
	\scalebox{0.8}{
		\begin{tabular}{c|c|c|c|c}
			\toprule
			\textbf{Learning Rate} & \textbf{Sample Size} & \textbf{Expected Cost} & \textbf{Macro F1} & \textbf{Cross-Entropy Loss} \\
			\midrule
			\multirow{3}{*}{0.0001} 
			& 2595  & 199930.644 ($\pm$ 12345.880) & 0.557 ($\pm$ 0.030) & 0.054 ($\pm$ 0.005) \\
			& 5190  & 166067.178 ($\pm$  6721.855) & 0.600 ($\pm$ 0.019) & 0.043 ($\pm$ 0.001) \\
			& 25946 & 152893.610 ($\pm$  6508.349) & 0.606 ($\pm$ 0.008) & 0.040 ($\pm$ 0.000) \\
			\midrule
			\multirow{3}{*}{0.001} 
			& 2595  & 126633.267 ($\pm$ 29351.125) & 0.702 ($\pm$ 0.035) & 0.071 ($\pm$ 0.016) \\
			& 5190  &  72263.458 ($\pm$  4378.611) & 0.838 ($\pm$ 0.003) & 0.026 ($\pm$ 0.001) \\
			& 25946 &  66377.267 ($\pm$   357.897) & 0.859 ($\pm$ 0.003) & 0.020 ($\pm$ 0.000) \\
			\midrule
			\multirow{3}{*}{0.005} 
			& 2595  & 130644.749 ($\pm$ 39320.791) & 0.684 ($\pm$ 0.053) & 0.065 ($\pm$ 0.017) \\
			& 5190  &  72251.576 ($\pm$  3609.323) & 0.832 ($\pm$ 0.000) & 0.027 ($\pm$ 0.001) \\
			& 25946 &  66648.000 ($\pm$   632.578) & 0.859 ($\pm$ 0.004) & 0.020 ($\pm$ 0.000) \\
			\bottomrule
	\end{tabular}}
\end{table}

For completeness, we report the full set of domain-specific evaluation metrics across all investigated values of $K$. Specifically, we evaluated Precision@K, Recall@K and Expected Cost@K for $K\in\{0.1, 0.2, 0.5, 1\}$ and partial PR-AUC for recall thresholds $r\in\{0.6, 0.7, 0.8, 0.9\}$ for the.

\cref{tab:appendix_costK} presents the expected cost for all models and labeled training sample sizes across different values of $K$, directly reflecting the financial impact of false predictions. This metric provides the most application-relevant evaluation.

\begin{table}[H]
	\centering
	\caption{Expected Cost@K for different models, training sizes and values of $K$. Values reported as mean ($\pm$ std.) in thousands. Bold values indicate the best result per row.}
	\label{tab:appendix_costK}
	\scalebox{0.7}{
	\begin{tabular}{c|c|cccccccc}
		\toprule
		\textbf{K} & \textbf{$N_l$} & \textbf{NB} & \textbf{L} & \textbf{L-SSL} & \textbf{KNN} & \textbf{SVM} & \textbf{RF} & \textbf{FN} & \textbf{Our model} \\
		\midrule
		\multirow{5}{*}{0.1} 
		& 2595  & 339.8 ($\pm$2.1) & 298.6 ($\pm$10.9) & 278.7 ($\pm$32.9) & 326.3 ($\pm$5.7) & 218.0 ($\pm$20.8) & 289.5 ($\pm$15.7) & 244.2 ($\pm$37.1) & \textbf{194.1 ($\pm$1.6)} \\
		& 3893  & 339.6 ($\pm$7.5) & 277.5 ($\pm$16.3) & 239.6 ($\pm$29.6) & 311.0 ($\pm$4.5) & 209.2 ($\pm$21.1) & 279.6 ($\pm$15.6) & 239.0 ($\pm$11.7) & \textbf{192.9 ($\pm$1.4)} \\
		& 5190  & 339.3 ($\pm$4.3) & 256.3 ($\pm$10.9) & 234.7 ($\pm$22.3) & 305.4 ($\pm$7.0) & 197.2 ($\pm$5.3)  & 288.3 ($\pm$11.2) & 227.9 ($\pm$26.7) & \textbf{198.6 ($\pm$9.6)} \\
		& 12973 & 336.3 ($\pm$5.8) & 216.1 ($\pm$2.9)  & 209.9 ($\pm$4.8)  & 301.9 ($\pm$5.4) & 200.7 ($\pm$14.8) & 284.8 ($\pm$16.2) & 218.4 ($\pm$22.3) & \textbf{197.1 ($\pm$6.4)} \\
		& 25946 & 337.3 ($\pm$7.1) & 203.3 ($\pm$4.0)  & 206.3 ($\pm$1.4)  & 302.5 ($\pm$8.8) & 209.6 ($\pm$12.4) & 280.3 ($\pm$18.7) & 193.1 ($\pm$1.9)  & \textbf{194.1 ($\pm$1.6)} \\
		\midrule
		\multirow{5}{*}{0.2} 
		& 2595  & 331.0 ($\pm$4.7)  & 275.1 ($\pm$15.9) & 248.0 ($\pm$33.6) & 310.8 ($\pm$10.6) & 182.5 ($\pm$17.9) & 240.3 ($\pm$21.4) & 184.7 ($\pm$37.1) & \textbf{152.1 ($\pm$3.8)} \\
		& 3893  & 333.7 ($\pm$10.9) & 252.3 ($\pm$16.2) & 216.9 ($\pm$22.9) & 297.1 ($\pm$3.1)  & 168.5 ($\pm$20.9) & 228.7 ($\pm$13.7) & 159.1 ($\pm$13.9) & \textbf{149.7 ($\pm$4.6)} \\
		& 5190  & 331.7 ($\pm$4.3)  & 231.8 ($\pm$20.3) & 210.9 ($\pm$15.2) & 294.8 ($\pm$4.7)  & 159.3 ($\pm$6.5)  & 237.9 ($\pm$12.8) & 152.9 ($\pm$7.2)  & \textbf{151.0 ($\pm$5.9)} \\
		& 12973 & 328.9 ($\pm$7.8)  & 195.4 ($\pm$3.2)  & 184.9 ($\pm$6.1)  & 266.5 ($\pm$5.1)  & 149.8 ($\pm$8.0)  & 226.7 ($\pm$19.0) & 161.1 ($\pm$13.4) & \textbf{150.9 ($\pm$5.3)} \\
		& 25946 & 330.4 ($\pm$7.7)  & 176.1 ($\pm$3.2)  & 171.1 ($\pm$2.2)  & 249.6 ($\pm$12.2) & 146.6 ($\pm$3.8)  & 221.7 ($\pm$15.8) & 151.3 ($\pm$2.1)  & \textbf{149.9 ($\pm$2.2)} \\
		\midrule
		\multirow{5}{*}{0.5} 
		& 2595  & 308.7 ($\pm$9.0)  & 226.9 ($\pm$16.6) & 193.7 ($\pm$32.3) & 285.6 ($\pm$9.1)  & 128.9 ($\pm$25.9) & 146.6 ($\pm$27.0) & 96.0 ($\pm$24.5)  & \textbf{79.8 ($\pm$9.5)} \\
		& 3893  & 308.3 ($\pm$15.6) & 205.2 ($\pm$14.2) & 170.2 ($\pm$11.0) & 273.7 ($\pm$9.4)  & 107.4 ($\pm$27.2) & 130.7 ($\pm$18.0) & 81.9 ($\pm$8.9)  & \textbf{77.8 ($\pm$9.8)} \\
		& 5190  & 310.3 ($\pm$5.5)  & 191.1 ($\pm$11.7) & 162.9 ($\pm$6.2)  & 267.1 ($\pm$10.2) & 88.6 ($\pm$9.4)  & 127.2 ($\pm$21.1) & 78.5 ($\pm$6.8)  & \textbf{73.6 ($\pm$3.4)} \\
		& 12973 & 304.4 ($\pm$8.4)  & 159.4 ($\pm$6.9)  & 142.8 ($\pm$3.3)  & 237.6 ($\pm$10.1) & 74.9 ($\pm$8.6)  & 110.4 ($\pm$10.3) & 72.0 ($\pm$4.2)  & \textbf{68.4 ($\pm$1.1)} \\
		& 25946 & 309.8 ($\pm$8.1)  & 137.2 ($\pm$5.6)  & 127.5 ($\pm$4.8)  & 205.9 ($\pm$12.2) & 72.4 ($\pm$2.9)  & 103.5 ($\pm$8.0)  & 68.0 ($\pm$1.1)  & \textbf{66.5 ($\pm$0.6)} \\
		\midrule
		\multirow{5}{*}{1.0} 
		& 2595  & 271.4 ($\pm$20.0) & 189.9 ($\pm$10.7) & 168.1 ($\pm$29.8) & 249.0 ($\pm$15.5) & 74.7 ($\pm$15.7) & 59.0 ($\pm$29.8) & 45.4 ($\pm$9.0)  & \textbf{43.5 ($\pm$10.6)} \\
		& 3893  & 263.8 ($\pm$20.5) & 169.4 ($\pm$18.3) & 141.8 ($\pm$12.0) & 228.2 ($\pm$17.0) & 57.2 ($\pm$15.9) & 57.6 ($\pm$26.1) & 40.1 ($\pm$6.1)  & \textbf{38.0 ($\pm$8.7)} \\
		& 5190  & 270.6 ($\pm$11.0) & 157.5 ($\pm$10.5) & 137.0 ($\pm$7.3)  & 222.7 ($\pm$17.0) & 43.5 ($\pm$6.6)  & 53.0 ($\pm$20.0) & 38.1 ($\pm$7.6)  & \textbf{32.2 ($\pm$6.1)} \\
		& 12973 & 259.9 ($\pm$19.9) & 124.6 ($\pm$5.6)  & 108.8 ($\pm$4.0)  & 184.3 ($\pm$11.6) & 31.7 ($\pm$4.4)  & 34.0 ($\pm$8.5)  & 26.6 ($\pm$2.1)  & \textbf{23.3 ($\pm$1.7)} \\
		& 25946 & 261.3 ($\pm$14.3) & 102.2 ($\pm$3.0)  & 88.1 ($\pm$5.9)   & 148.4 ($\pm$10.3) & 27.8 ($\pm$2.1)  & 28.9 ($\pm$6.6)  & \textbf{22.0 ($\pm$1.6)} & 22.2 ($\pm$1.4) \\
		\bottomrule
	\end{tabular}}
\end{table}

\cref{tab:appendix_precisionK} reports Precision@K across the same setting. It shows how effective each model identifies true frauds within the top-$K$ \% ranked transactions. This is especially critical to minimize unnecessary investigations.

\begin{table}[H]
	\centering
	\caption{Precision@K for different models, training sizes and values of $K$. Values reported as mean ($\pm$ std.). Bold values indicate the best result per row.}
	\label{tab:appendix_precisionK}
	\scalebox{0.65}{
	\begin{tabular}{c|c|cccccccc}
		\toprule
		\textbf{K} & \textbf{$N_l$} & \textbf{NB} & \textbf{L} & \textbf{L-SSL} & \textbf{KNN} & \textbf{SVM} & \textbf{RF} & \textbf{FN} & \textbf{Our model} \\
		\midrule
		\multirow{5}{*}{0.1} 
		& 2595  & 0.224 ($\pm$0.085) & 0.345 ($\pm$0.065) & 0.559 ($\pm$0.111) & 0.355 ($\pm$0.103) & 0.721 ($\pm$0.166) & 0.934 ($\pm$0.037) & 0.924 ($\pm$0.040) & \textbf{0.948 ($\pm$0.024)} \\
		& 3893  & 0.197 ($\pm$0.182) & 0.431 ($\pm$0.068) & 0.693 ($\pm$0.123) & 0.531 ($\pm$0.091) & 0.800 ($\pm$0.177) & \textbf{0.962 ($\pm$0.026)} & 0.955 ($\pm$0.029) & 0.948 ($\pm$0.012) \\
		& 5190  & 0.255 ($\pm$0.097) & 0.545 ($\pm$0.065) & 0.707 ($\pm$0.117) & 0.579 ($\pm$0.081) & 0.886 ($\pm$0.081) & 0.941 ($\pm$0.031) & \textbf{0.962 ($\pm$0.022)} & 0.948 ($\pm$0.012) \\
		& 12973 & 0.234 ($\pm$0.038) & 0.745 ($\pm$0.031) & 0.745 ($\pm$0.048) & 0.772 ($\pm$0.079) & 0.910 ($\pm$0.067) & \textbf{0.976 ($\pm$0.009)} & 0.969 ($\pm$0.014) & 0.962 ($\pm$0.008) \\
		& 25946 & 0.238 ($\pm$0.104) & 0.834 ($\pm$0.043) & 0.772 ($\pm$0.019) & 0.800 ($\pm$0.059) & 0.928 ($\pm$0.066) & 0.962 ($\pm$0.022) & \textbf{0.976 ($\pm$0.009)} & 0.962 ($\pm$0.008) \\
		\midrule
		\multirow{5}{*}{0.2} 
		& 2595  & 0.212 ($\pm$0.071) & 0.383 ($\pm$0.082) & 0.493 ($\pm$0.055) & 0.353 ($\pm$0.085) & 0.679 ($\pm$0.135) & \textbf{0.912 ($\pm$0.052)} & 0.907 ($\pm$0.035) & 0.929 ($\pm$0.040) \\
		& 3893  & 0.183 ($\pm$0.132) & 0.416 ($\pm$0.113) & 0.541 ($\pm$0.063) & 0.412 ($\pm$0.082) & 0.783 ($\pm$0.161) & 0.933 ($\pm$0.014) & 0.938 ($\pm$0.015) & \textbf{0.953 ($\pm$0.014)} \\
		& 5190  & 0.252 ($\pm$0.050) & 0.479 ($\pm$0.120) & 0.560 ($\pm$0.035) & 0.407 ($\pm$0.077) & 0.852 ($\pm$0.043) & 0.936 ($\pm$0.018) & 0.953 ($\pm$0.023) & \textbf{0.952 ($\pm$0.012)} \\
		& 12973 & 0.224 ($\pm$0.032) & 0.653 ($\pm$0.047) & 0.700 ($\pm$0.027) & 0.700 ($\pm$0.036) & 0.921 ($\pm$0.064) & 0.952 ($\pm$0.013) & 0.959 ($\pm$0.014) & \textbf{0.953 ($\pm$0.010)} \\
		& 25946 & 0.214 ($\pm$0.087) & 0.779 ($\pm$0.012) & 0.778 ($\pm$0.022) & 0.810 ($\pm$0.066) & 0.945 ($\pm$0.042) & 0.964 ($\pm$0.019) & 0.957 ($\pm$0.006) & \textbf{0.959 ($\pm$0.007)} \\
		\midrule
		\multirow{5}{*}{0.5} 
		& 2595  & 0.201 ($\pm$0.027) & 0.331 ($\pm$0.036) & 0.394 ($\pm$0.045) & 0.266 ($\pm$0.033) & 0.594 ($\pm$0.123) & 0.819 ($\pm$0.079) & 0.833 ($\pm$0.055) & \textbf{0.883 ($\pm$0.044)} \\
		& 3893  & 0.216 ($\pm$0.075) & 0.366 ($\pm$0.027) & 0.441 ($\pm$0.029) & 0.289 ($\pm$0.056) & 0.709 ($\pm$0.138) & 0.852 ($\pm$0.051) & 0.877 ($\pm$0.038) & \textbf{0.896 ($\pm$0.044)} \\
		& 5190  & 0.225 ($\pm$0.042) & 0.392 ($\pm$0.030) & 0.469 ($\pm$0.010) & 0.305 ($\pm$0.073) & 0.802 ($\pm$0.046) & 0.880 ($\pm$0.036) & 0.910 ($\pm$0.014) & \textbf{0.919 ($\pm$0.020)} \\
		& 12973 & 0.216 ($\pm$0.018) & 0.490 ($\pm$0.032) & 0.541 ($\pm$0.017) & 0.442 ($\pm$0.035) & 0.889 ($\pm$0.058) & 0.920 ($\pm$0.013) & \textbf{0.945 ($\pm$0.008)} & 0.944 ($\pm$0.005) \\
		& 25946 & 0.188 ($\pm$0.031) & 0.583 ($\pm$0.034) & 0.621 ($\pm$0.024) & 0.549 ($\pm$0.048) & 0.898 ($\pm$0.024) & 0.938 ($\pm$0.012) & \textbf{0.958 ($\pm$0.009)} & 0.952 ($\pm$0.008) \\
		\midrule
		\multirow{5}{*}{1.0} 
		& 2595  & 0.207 ($\pm$0.038) & 0.271 ($\pm$0.022) & 0.280 ($\pm$0.030) & 0.208 ($\pm$0.033) & 0.508 ($\pm$0.057) & \textbf{0.669 ($\pm$0.060)} & 0.649 ($\pm$0.049) & 0.655 ($\pm$0.048) \\
		& 3893  & 0.218 ($\pm$0.043) & 0.298 ($\pm$0.035) & 0.319 ($\pm$0.027) & 0.251 ($\pm$0.023) & 0.573 ($\pm$0.072) & \textbf{0.681 ($\pm$0.050)} & 0.677 ($\pm$0.028) & 0.684 ($\pm$0.039) \\
		& 5190  & 0.211 ($\pm$0.020) & 0.314 ($\pm$0.025) & 0.338 ($\pm$0.012) & 0.261 ($\pm$0.030) & 0.636 ($\pm$0.031) & 0.701 ($\pm$0.043) & 0.700 ($\pm$0.028) & \textbf{0.715 ($\pm$0.032)} \\
		& 12973 & 0.214 ($\pm$0.018) & 0.383 ($\pm$0.016) & 0.416 ($\pm$0.015) & 0.359 ($\pm$0.021) & 0.698 ($\pm$0.022) & 0.751 ($\pm$0.018) & 0.756 ($\pm$0.017) & \textbf{0.764 ($\pm$0.010)} \\
		& 25946 & 0.214 ($\pm$0.034) & 0.447 ($\pm$0.009) & 0.472 ($\pm$0.019) & 0.440 ($\pm$0.022) & 0.721 ($\pm$0.014) & 0.766 ($\pm$0.016) & \textbf{0.779 ($\pm$0.013)} & 0.770 ($\pm$0.007) \\
		\bottomrule
	\end{tabular}}
\end{table}

\cref{tab:appendix_recallK} provides Recall@K results, reflecting the share of frauds captured within the top-$K$ \%.

\begin{table}[ht]
	\centering
	\caption{Recall@K for different models, training sizes and values of $K$. Values reported as mean ($\pm$ std.). Bold values indicate the best result per row.}
	\label{tab:appendix_recallK}
	\scalebox{0.65}{
	\begin{tabular}{c|c|cccccccc}
		\toprule
		\textbf{K} & \textbf{$N_l$} & \textbf{NB} & \textbf{L} & \textbf{L-SSL} & \textbf{KNN} & \textbf{SVM} & \textbf{RF} & \textbf{FN} & \textbf{Our model} \\
		\midrule
		\multirow{5}{*}{0.1} 
		& 2595  & 0.020 ($\pm$0.008) & 0.031 ($\pm$0.006) & 0.051 ($\pm$0.010) & 0.032 ($\pm$0.009) & 0.065 ($\pm$0.015) & 0.085 ($\pm$0.003) & 0.084 ($\pm$0.004) & \textbf{0.086 ($\pm$0.002)} \\
		& 3893  & 0.018 ($\pm$0.017) & 0.039 ($\pm$0.006) & 0.063 ($\pm$0.011) & 0.048 ($\pm$0.008) & 0.073 ($\pm$0.016) & \textbf{0.087 ($\pm$0.002)} & 0.087 ($\pm$0.003) & 0.086 ($\pm$0.001) \\
		& 5190  & 0.023 ($\pm$0.009) & 0.049 ($\pm$0.006) & 0.064 ($\pm$0.011) & 0.053 ($\pm$0.007) & 0.080 ($\pm$0.007) & 0.085 ($\pm$0.003) & \textbf{0.087 ($\pm$0.002)} & 0.086 ($\pm$0.001) \\
		& 12973 & 0.021 ($\pm$0.003) & 0.068 ($\pm$0.003) & 0.068 ($\pm$0.004) & 0.070 ($\pm$0.007) & 0.083 ($\pm$0.006) & \textbf{0.089 ($\pm$0.001)} & 0.088 ($\pm$0.001) & 0.087 ($\pm$0.001) \\
		& 25946 & 0.022 ($\pm$0.009) & 0.076 ($\pm$0.004) & 0.070 ($\pm$0.002) & 0.073 ($\pm$0.005) & 0.084 ($\pm$0.006) & 0.087 ($\pm$0.002) & \textbf{0.089 ($\pm$0.001)} & 0.087 ($\pm$0.001) \\
		\midrule
		\multirow{5}{*}{0.2} 
		& 2595  & 0.038 ($\pm$0.013) & 0.069 ($\pm$0.015) & 0.090 ($\pm$0.010) & 0.064 ($\pm$0.015) & 0.123 ($\pm$0.025) & 0.166 ($\pm$0.009) & 0.165 ($\pm$0.006) & \textbf{0.169 ($\pm$0.007)} \\
		& 3893  & 0.033 ($\pm$0.024) & 0.075 ($\pm$0.021) & 0.098 ($\pm$0.012) & 0.075 ($\pm$0.015) & 0.142 ($\pm$0.029) & 0.169 ($\pm$0.003) & 0.170 ($\pm$0.003) & \textbf{0.173 ($\pm$0.003)} \\
		& 5190  & 0.046 ($\pm$0.009) & 0.087 ($\pm$0.022) & 0.102 ($\pm$0.006) & 0.074 ($\pm$0.014) & 0.155 ($\pm$0.008) & 0.170 ($\pm$0.003) & \textbf{0.173 ($\pm$0.004)} & 0.173 ($\pm$0.002) \\
		& 12973 & 0.041 ($\pm$0.006) & 0.119 ($\pm$0.008) & 0.127 ($\pm$0.005) & 0.127 ($\pm$0.007) & 0.167 ($\pm$0.012) & 0.173 ($\pm$0.002) & \textbf{0.174 ($\pm$0.003)} & 0.173 ($\pm$0.002) \\
		& 25946 & 0.039 ($\pm$0.016) & 0.141 ($\pm$0.002) & 0.141 ($\pm$0.004) & 0.147 ($\pm$0.012) & 0.172 ($\pm$0.008) & \textbf{0.175 ($\pm$0.003)} & 0.174 ($\pm$0.001) & 0.174 ($\pm$0.001) \\
		\midrule
		\multirow{5}{*}{0.5} 
		& 2595  & 0.091 ($\pm$0.012) & 0.150 ($\pm$0.016) & 0.178 ($\pm$0.020) & 0.120 ($\pm$0.015) & 0.269 ($\pm$0.056) & 0.370 ($\pm$0.036) & 0.377 ($\pm$0.025) & \textbf{0.399 ($\pm$0.020)} \\
		& 3893  & 0.098 ($\pm$0.034) & 0.166 ($\pm$0.012) & 0.199 ($\pm$0.013) & 0.131 ($\pm$0.025) & 0.321 ($\pm$0.062) & 0.385 ($\pm$0.023) & 0.397 ($\pm$0.017) & \textbf{0.405 ($\pm$0.020)} \\
		& 5190  & 0.102 ($\pm$0.019) & 0.177 ($\pm$0.014) & 0.212 ($\pm$0.005) & 0.138 ($\pm$0.033) & 0.363 ($\pm$0.021) & 0.398 ($\pm$0.016) & 0.412 ($\pm$0.006) & \textbf{0.416 ($\pm$0.009)} \\
		& 12973 & 0.098 ($\pm$0.008) & 0.222 ($\pm$0.014) & 0.245 ($\pm$0.008) & 0.200 ($\pm$0.016) & 0.402 ($\pm$0.026) & 0.416 ($\pm$0.006) & \textbf{0.428 ($\pm$0.003)} & 0.427 ($\pm$0.002) \\
		& 25946 & 0.085 ($\pm$0.014) & 0.264 ($\pm$0.015) & 0.281 ($\pm$0.011) & 0.248 ($\pm$0.022) & 0.406 ($\pm$0.011) & 0.424 ($\pm$0.005) & \textbf{0.433 ($\pm$0.004)} & 0.431 ($\pm$0.004) \\
		\midrule
		\multirow{5}{*}{1.0} 
		& 2595  & 0.187 ($\pm$0.034) & 0.245 ($\pm$0.020) & 0.253 ($\pm$0.027) & 0.188 ($\pm$0.029) & 0.459 ($\pm$0.051) & 0.604 ($\pm$0.054) & 0.586 ($\pm$0.044) & \textbf{0.591 ($\pm$0.043)} \\
		& 3893  & 0.197 ($\pm$0.039) & 0.269 ($\pm$0.031) & 0.288 ($\pm$0.024) & 0.226 ($\pm$0.021) & 0.518 ($\pm$0.065) & 0.615 ($\pm$0.045) & 0.611 ($\pm$0.025) & \textbf{0.617 ($\pm$0.035)} \\
		& 5190  & 0.191 ($\pm$0.018) & 0.284 ($\pm$0.022) & 0.305 ($\pm$0.011) & 0.235 ($\pm$0.027) & 0.575 ($\pm$0.028) & 0.633 ($\pm$0.039) & 0.632 ($\pm$0.025) & \textbf{0.645 ($\pm$0.029)} \\
		& 12973 & 0.193 ($\pm$0.016) & 0.346 ($\pm$0.014) & 0.375 ($\pm$0.014) & 0.324 ($\pm$0.019) & 0.631 ($\pm$0.020) & 0.679 ($\pm$0.016) & 0.683 ($\pm$0.015) & \textbf{0.690 ($\pm$0.009)} \\
		& 25946 & 0.193 ($\pm$0.031) & 0.403 ($\pm$0.008) & 0.427 ($\pm$0.017) & 0.397 ($\pm$0.020) & 0.651 ($\pm$0.012) & 0.691 ($\pm$0.014) & \textbf{0.704 ($\pm$0.012)} & 0.695 ($\pm$0.006) \\
		\bottomrule
	\end{tabular}}
\end{table}

Finally, \cref{tab:appendix_prauck} shows the partial PR-AUC at different recall thresholds, investigation model performance in realistic operational regimes, where full recall is not achievable.

\begin{table}[H]
	\centering
	\caption{Partial PR-AUC comparison of models across different $K$ values and training sizes. Values reported as mean ($\pm$ std.). Bold values indicate the best result per row.}
	\label{tab:appendix_prauck}
	\scalebox{0.65}{
	\begin{tabular}{c|c|cccccccc}
		\toprule
		\textbf{K} & \textbf{$N_l$} & \textbf{NB} & \textbf{L} & \textbf{L-SSL} & \textbf{KNN} & \textbf{SVM} & \textbf{RF} & \textbf{FN} & \textbf{Our model} \\
		\midrule
		\multirow{5}{*}{0.5} 
		& 2595  & 0.377 ($\pm$0.000) & 0.133 ($\pm$0.012) & 0.163 ($\pm$0.022) & 0.254 ($\pm$0.027) & 0.316 ($\pm$0.064) & 0.434 ($\pm$0.034) & 0.441 ($\pm$0.022) & \textbf{0.455 ($\pm$0.020)} \\
		& 3893  & 0.377 ($\pm$0.000) & 0.161 ($\pm$0.015) & 0.201 ($\pm$0.027) & 0.276 ($\pm$0.011) & 0.370 ($\pm$0.082) & 0.451 ($\pm$0.017) & 0.457 ($\pm$0.014) & \textbf{0.466 ($\pm$0.013)} \\
		& 5190  & 0.377 ($\pm$0.000) & 0.180 ($\pm$0.017) & 0.214 ($\pm$0.017) & 0.288 ($\pm$0.016) & 0.418 ($\pm$0.017) & 0.457 ($\pm$0.013) & \textbf{0.470 ($\pm$0.004)} & 0.470 ($\pm$0.006) \\
		& 12973 & 0.377 ($\pm$0.000) & 0.256 ($\pm$0.015) & 0.280 ($\pm$0.014) & 0.336 ($\pm$0.013) & 0.456 ($\pm$0.028) & 0.472 ($\pm$0.004) & \textbf{0.480 ($\pm$0.003)} & 0.479 ($\pm$0.003) \\
		& 25946 & 0.377 ($\pm$0.000) & 0.320 ($\pm$0.010) & 0.328 ($\pm$0.010) & 0.370 ($\pm$0.014) & 0.463 ($\pm$0.017) & 0.477 ($\pm$0.005) & \textbf{0.483 ($\pm$0.002)} & 0.482 ($\pm$0.002) \\
		\midrule
		\multirow{5}{*}{0.6} 
		& 2595  & 0.422 ($\pm$0.000) & 0.141 ($\pm$0.012) & 0.174 ($\pm$0.025) & 0.315 ($\pm$0.026) & 0.360 ($\pm$0.072) & 0.505 ($\pm$0.043) & 0.510 ($\pm$0.030) & \textbf{0.527 ($\pm$0.029)} \\
		& 3893  & 0.423 ($\pm$0.001) & 0.169 ($\pm$0.015) & 0.214 ($\pm$0.026) & 0.338 ($\pm$0.012) & 0.428 ($\pm$0.094) & 0.524 ($\pm$0.025) & 0.533 ($\pm$0.019) & \textbf{0.544 ($\pm$0.021)} \\
		& 5190  & 0.423 ($\pm$0.001) & 0.189 ($\pm$0.017) & 0.230 ($\pm$0.016) & 0.350 ($\pm$0.017) & 0.485 ($\pm$0.023) & 0.534 ($\pm$0.020) & 0.552 ($\pm$0.005) & \textbf{0.553 ($\pm$0.008)} \\
		& 12973 & 0.423 ($\pm$0.001) & 0.274 ($\pm$0.016) & 0.305 ($\pm$0.013) & 0.404 ($\pm$0.013) & 0.536 ($\pm$0.033) & 0.559 ($\pm$0.005) & \textbf{0.570 ($\pm$0.004)} & 0.569 ($\pm$0.004) \\
		& 25946 & 0.423 ($\pm$0.000) & 0.348 ($\pm$0.012) & 0.362 ($\pm$0.012) & 0.442 ($\pm$0.015) & 0.545 ($\pm$0.020) & 0.567 ($\pm$0.006) & \textbf{0.576 ($\pm$0.002)} & 0.574 ($\pm$0.002) \\
		\midrule
		\multirow{5}{*}{0.7} 
		& 2595  & 0.458 ($\pm$0.001) & 0.147 ($\pm$0.013) & 0.183 ($\pm$0.025) & 0.375 ($\pm$0.026) & 0.395 ($\pm$0.077) & 0.569 ($\pm$0.051) & 0.560 ($\pm$0.043) & \textbf{0.582 ($\pm$0.038)} \\
		& 3893  & 0.459 ($\pm$0.001) & 0.176 ($\pm$0.016) & 0.224 ($\pm$0.027) & 0.399 ($\pm$0.012) & 0.471 ($\pm$0.099) & 0.590 ($\pm$0.032) & 0.590 ($\pm$0.027) & \textbf{0.606 ($\pm$0.027)} \\
		& 5190  & 0.459 ($\pm$0.001) & 0.197 ($\pm$0.017) & 0.240 ($\pm$0.015) & 0.413 ($\pm$0.018) & 0.538 ($\pm$0.028) & 0.602 ($\pm$0.028) & 0.616 ($\pm$0.013) & \textbf{0.624 ($\pm$0.013)} \\
		& 12973 & 0.459 ($\pm$0.001) & 0.283 ($\pm$0.017) & 0.318 ($\pm$0.013) & 0.472 ($\pm$0.014) & 0.603 ($\pm$0.038) & 0.637 ($\pm$0.007) & 0.649 ($\pm$0.007) & \textbf{0.651 ($\pm$0.006)} \\
		& 25946 & 0.459 ($\pm$0.000) & 0.363 ($\pm$0.014) & 0.383 ($\pm$0.012) & 0.513 ($\pm$0.016) & 0.618 ($\pm$0.021) & 0.649 ($\pm$0.007) & \textbf{0.660 ($\pm$0.003)} & 0.657 ($\pm$0.004) \\
		\midrule
		\multirow{5}{*}{0.8} 
		& 2595  & 0.479 ($\pm$0.005) & 0.152 ($\pm$0.012) & 0.189 ($\pm$0.024) & 0.435 ($\pm$0.025) & 0.411 ($\pm$0.078) & 0.622 ($\pm$0.059) & 0.581 ($\pm$0.055) & \textbf{0.610 ($\pm$0.042)} \\
		& 3893  & 0.482 ($\pm$0.005) & 0.182 ($\pm$0.017) & 0.231 ($\pm$0.026) & 0.461 ($\pm$0.013) & 0.492 ($\pm$0.101) & 0.642 ($\pm$0.040) & 0.619 ($\pm$0.036) & \textbf{0.642 ($\pm$0.031)} \\
		& 5190  & 0.483 ($\pm$0.003) & 0.204 ($\pm$0.017) & 0.248 ($\pm$0.015) & 0.476 ($\pm$0.019) & 0.569 ($\pm$0.030) & 0.655 ($\pm$0.036) & 0.652 ($\pm$0.022) & \textbf{0.672 ($\pm$0.018)} \\
		& 12973 & 0.485 ($\pm$0.002) & 0.291 ($\pm$0.017) & 0.327 ($\pm$0.013) & 0.539 ($\pm$0.015) & 0.647 ($\pm$0.040) & 0.703 ($\pm$0.009) & 0.710 ($\pm$0.011) & \textbf{0.714 ($\pm$0.008)} \\
		& 25946 & 0.483 ($\pm$0.003) & 0.373 ($\pm$0.015) & 0.395 ($\pm$0.013) & 0.584 ($\pm$0.017) & 0.669 ($\pm$0.022) & 0.717 ($\pm$0.009) & \textbf{0.728 ($\pm$0.006)} & 0.723 ($\pm$0.005) \\
		\bottomrule
	\end{tabular}}
\end{table}


	\addcontentsline{toc}{section}{References}
	\bibliographystyle{unsrtnat}
	\bibliography{bib}
\end{document}